\PassOptionsToPackage{table}{xcolor}
\documentclass{article}
\usepackage[T1]{fontenc}
\usepackage{iclr2027_conference,times}

\usepackage{amsmath,amsfonts,bm}

\def\eqref#1{equation~\ref{#1}}

\def\1{\bm{1}}

\DeclareMathAlphabet{\mathsfit}{\encodingdefault}{\sfdefault}{m}{sl}
\SetMathAlphabet{\mathsfit}{bold}{\encodingdefault}{\sfdefault}{bx}{n}

\usepackage{hyperref}
\usepackage{url}
\usepackage{booktabs}
\usepackage{multirow}
\usepackage{graphicx}
\graphicspath{{./}}
\usepackage{amsmath}
\usepackage{xcolor}
\usepackage{enumitem}
\usepackage{algorithm}   
\usepackage{algorithmic}
\usepackage{subcaption}
\usepackage{booktabs}
\usepackage{graphicx}
\usepackage{wrapfig}
\usepackage[table]{xcolor}
\definecolor{seedsbg}{HTML}{BFDFD2}

\newcommand{\seeds}{\textsc{Seeds}}

\newcommand{\jl}[1]{{\color{blue}\textbf{[jl:} #1\textbf{]}}}

\title{%
\protect\raisebox{-0.1em}{%
    \protect\includegraphics[height=0.9em]{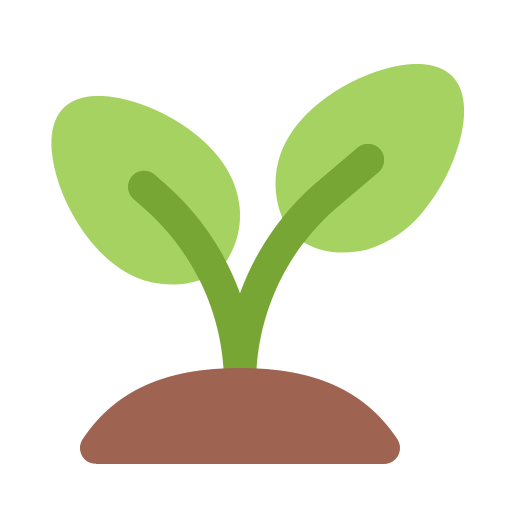}%
  }\hspace{0.25em}%
\seeds: Preview-Guided Evidence Admission for Search Agents}

\author{Jiajin Liu$^{1,2*}$, Chenxi Xu$^{2*}$, Dongzhe Fan$^{1,2}$, Yuanfu Sun$^{1,2}$, Qiaoyu Tan$^{2}$ \\
$^{1}$New York University,
$^{2}$New York University Shanghai\\
\texttt{\{jiajinliu, qiaoyu.tan\}@nyu.edu}}

\iclrfinalcopy 
\begin{document}

\maketitle
\lhead{Preprint}

\begin{abstract}
Search agents extend large reasoning models with multi-round access to external knowledge, yet most existing systems largely conflate retrieving evidence with consuming it. 
Retrieved documents are commonly incorporated according to fixed or rank-based policies, despite the fact that evidence utility is conditioned on the agent's current reasoning state. Consequently, a lower-ranked candidate may be critical for the current step, while a higher-ranked candidate may provide redundant, irrelevant, or misleading information.
We formulate this overlooked problem as \textbf{evidence admission}, which asks, at each search step, which retrieved candidates should be allowed to enter the agent’s evolving reasoning context. To enable such selective control, we introduce \textbf{SEEDS (Selective Evidence Entry During Search)}, a preview-guided evidence admission framework for search agents. SEEDS first exposes lightweight previews of retrieved candidates and then lets the agent jointly determine which candidates and how many should be fully inspected, conditioned on its current reasoning state. Unlike fixed or adaptive prefix policies, SEEDS can select arbitrary subsets with flexible cardinality, and it can be instantiated directly with an existing reasoning agent or further optimized through learning. Across nine knowledge-intensive QA benchmarks under both local corpus and live web search, \seeds{} consistently improves search effectiveness while
substantially reducing full-evidence consumption. Experimental analyses show that \seeds{} more effectively selects useful candidates beyond the corresponding retrieval prefix. These
results establish evidence admission as a complementary dimension of agentic
search, separating what information is retrieved from what information is
ultimately allowed to shape reasoning. 

\end{abstract}

\section{Introduction}
\label{sec:intro}
Large Reasoning Models (LRMs) increasingly rely on external search to access up-to-date and task-relevant knowledge for solving knowledge-intensive problems. A standard approach is retrieval-augmented generation (RAG), which typically retrieves relevant information before generation and incorporates the retrieved evidence into the LRM input context~\citep{fan2024surveyragmeetingllms,gao2024ragsurvey,sharma2025ragsurvey}. However, complex questions often require iterative information seeking because the information needed by the model evolves as reasoning proceeds and cannot always be determined from the initial query alone~\citep{processbench,sorokin2026q}.

Search agents address this challenge by interleaving reasoning and retrieval over multiple rounds. At each round, an agent can formulate a new query based on its current reasoning state, retrieve additional information, and use the retrieved evidence to guide subsequent reasoning and search~\citep{jin2025searchr1,li2025searcho1,wei2026agentic,Li2025WebThinker}. This interaction creates an important distinction that is often overlooked: \emph{retrieving evidence does not necessarily mean that the evidence should be consumed by the agent}. Once retrieved information enters the reasoning trajectory, it can influence not only the current inference step, but also future search queries, subsequent reasoning decisions, and ultimately the final answer. Therefore, an effective search agent must decide not only what information to retrieve, but also which retrieved information should be allowed to influence its evolving reasoning process.


\begin{wrapfigure}{r}{0.36\columnwidth}
    \centering
    \includegraphics[width=\linewidth]{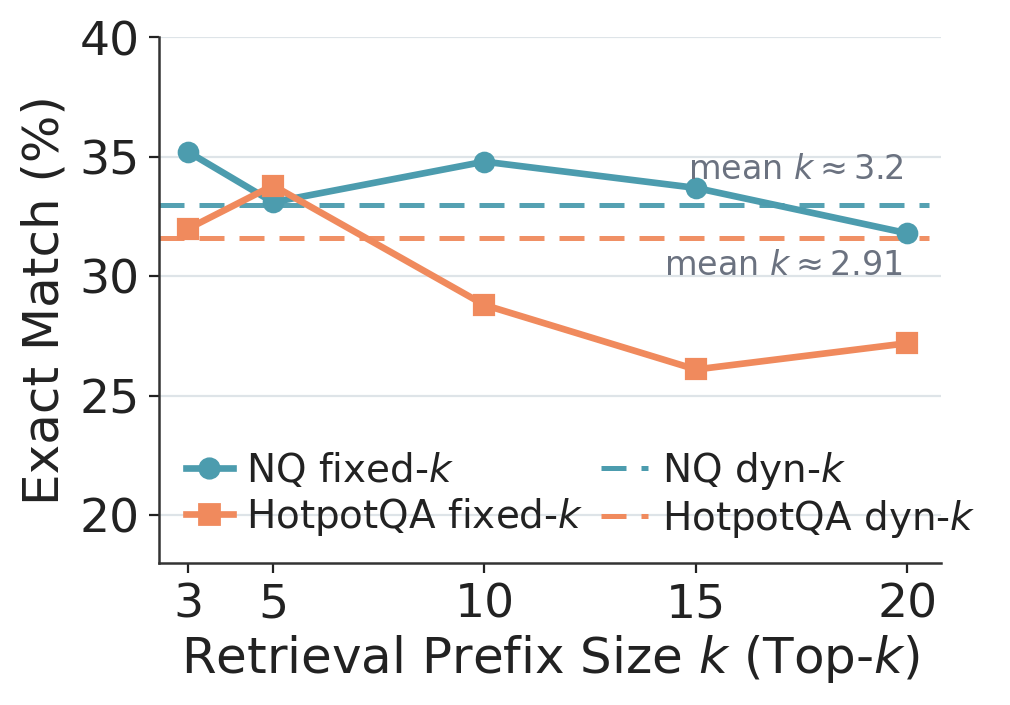}
    \vspace{-1em}
    \caption{Limitation of retrieval-rank prefix selection in multi-round search-augmented reasoning on QA tasks.
    Using a ReAct-style search agent with QwQ-32B, fixed-$k$ is compared with dyn-$k$, where the search agent adaptively selects the prefix size at each search round.
    }
    \label{fig:motivation}
    \vspace{-0.8em}
\end{wrapfigure}

We refer to this decision as \emph{evidence admission}. Evidence admission asks, at each search step, which retrieved candidates 
should enter the agent's reasoning context. This decision is inherently dependent on the current reasoning state. A document that provides essential information at one step may become redundant after related evidence has already been acquired, while a lower-ranked candidate may contain complementary information that is critical for resolving the agent's current uncertainty. Retrieval rank alone is therefore insufficient for determining the utility of evidence during reasoning~\citep{2023ircot,lee2025setr,2025mainrag,meng2025dps}. Existing search pipelines nevertheless commonly admit a fixed number of highly ranked results or restrict adaptive selection to a prefix of the retrieval ranking~\citep{jin2025searchr1,sharifymoghaddam2026rerank}. 
As illustrated in Fig.~\ref{fig:motivation}, the optimal retrieval-prefix size
varies across datasets, making it difficult to determine a suitable amount of evidence in advance. Allowing the model to adapt the prefix size at each search
round still falls short of the best fixed setting. This suggests that controlling evidence quantity alone is insufficient:
too little evidence may omit necessary information, while too much may introduce redundancy or distraction, and \emph{which} documents are admitted is equally important.

Selective evidence admission introduces a key challenge that the agent must assess the utility of retrieved document candidates \emph{before} consuming their full contents. Inspecting every candidate defeats the purpose of selective admission, while retrieval scores alone provide limited information about reasoning-time utility. We therefore introduce an intermediate representation, termed an \emph{evidence preview}, that provides a lightweight view of each candidate before full inspection. The preview is retrieval-environment agnostic: for local corpora, it can be instantiated as document identity together with a compact query-relevant excerpt, while for live web search, the title and snippet of results from search engines naturally serve this role. Evidence previews thus provide a low-cost interface between retrieval and full evidence consumption.

To enable such selective control, we introduce \seeds{} (\textbf{S}elective \textbf{E}vidence \textbf{E}ntry \textbf{D}uring \textbf{S}earch), a preview-guided evidence admission framework for search agents. After each retrieval step, \seeds{} first exposes lightweight previews of a candidate evidence pool. Conditioned on these previews and the agent's current reasoning state, the agent determines which candidates should be fully inspected and how many are necessary. Unlike fixed or adaptive prefix strategies, \seeds{} can admit an arbitrary subset of the retrieved candidates with flexible cardinality. Only the admitted candidates are expanded into detailed evidence and incorporated into the subsequent reasoning trajectory. The process can then repeat as the agent continues reasoning and searching. Importantly, \seeds{} is a general interaction framework rather than a specific learning algorithm. It can be instantiated directly with an existing reasoning agent, while supervised or reinforcement learning can further optimize the admission policy when training is available. We evaluate \seeds{} across nine knowledge-intensive QA benchmarks under both local corpus and live web search settings, using multiple search-agent paradigms. \seeds{} improves answer quality while reducing full-evidence consumption, and controlled analyses show that it more effectively admits useful candidates beyond the retrieval prefix. These results demonstrate that gains come from improving both \emph{which} evidence is admitted and \emph{how much} is consumed.
Our major \textbf{contributions} are summarized as follows:
\vspace{-1em}

\begin{itemize}[noitemsep,leftmargin=*]
    \item We formulate \emph{evidence admission} as a distinct decision problem in agentic search. This formulation separates evidence availability after retrieval from the decision of which evidence should enter and influence the evolving reasoning trajectory.

    \item We introduce \seeds{}, a general preview-guided evidence admission framework that allows a search agent to jointly determine which retrieved candidates to inspect and how many are needed at each reasoning step. Unlike rank-prefix strategies, \seeds{} supports flexible, non-prefix evidence selection conditioned on the current reasoning state.

    \item Across nine knowledge-intensive benchmarks with local and live search, we show that \seeds{} improves search effectiveness while reducing full-evidence consumption. Experimental analyses further demonstrate more effective evidence selection and show that \seeds{} can benefit from additional training without relying on learning for its basic operation.
    
\end{itemize}

\section{Related Work}
\label{sec:related}

Our work is most closely related to two lines of research.

\noindent{\textbf{Adaptive Retrieval in Retrieval-Augmented Generation.}}
Conventional retrieval-augmented generation (RAG) often provides language models with a fixed number of top-ranked passages, despite variation in the amount and composition of evidence required across queries. Self-RAG~\citep{asai2024selfrag} makes retrieval conditional on the model's generation needs, while Adaptive-$k$~\citep{taguchi2025dynamick} predicts a query-specific number of passages based on query--passage semantic similarity scores. DynamicRAG~\citep{sun2025dynamicrag} adapts passage ordering and set size using generation feedback. SETR~\citep{lee2025setr} and DPS~\citep{meng2025dps} move beyond individual passage relevance to select variable-size evidence sets that account for relationships among passages. MAIN-RAG~\citep{2025mainrag} filters candidates using multi-agent scoring and an adaptive threshold. 
These methods demonstrate the importance of adapting retrieved context rather than relying on a globally fixed number of passages. However, they primarily adapt evidence selection for a given query and retrieved candidate set. In contrast, \seeds{} considers evidence selection within a multi-round reasoning trajectory, where the selection decision is conditioned on the agent's evolving reasoning state.


\noindent{\textbf{Multi-Round Retrieval and Search Agents.}}
A complementary line of work integrates retrieval into multi-round search-augmented reasoning. IRCoT~\citep{2023ircot} interleaves reasoning with retrieval, while FLARE~\citep{jiang2023flare} and DRAGIN~\citep{su2024dragin} adapt when retrieval is invoked or how it is guided by the current generation state. More recent systems such as Search-o1~\citep{li2025searcho1}, Search-R1~\citep{jin2025searchr1}, and ReasonRAG~\citep{zhang2025process} optimize multi-round search behavior, while InfoAgent~\citep{zhang2025infoagent} and GRASP~\citep{gandhi2026grasp} provide richer mechanisms for browsing and inspecting retrieved information. These methods primarily address when and how to search, or how retrieved items are inspected. In contrast, \seeds{} introduces an explicit admission step between retrieval and evidence consumption. Given lightweight previews of multiple candidates, the agent jointly determines which candidates to inspect and how many are needed, conditioned on its current reasoning state. This enables flexible subset selection beyond fixed or adaptive retrieval prefixes.




\section{Methodology}
\label{sec:method}

\begin{figure}[t]
    \centering
    \includegraphics[width=\linewidth]{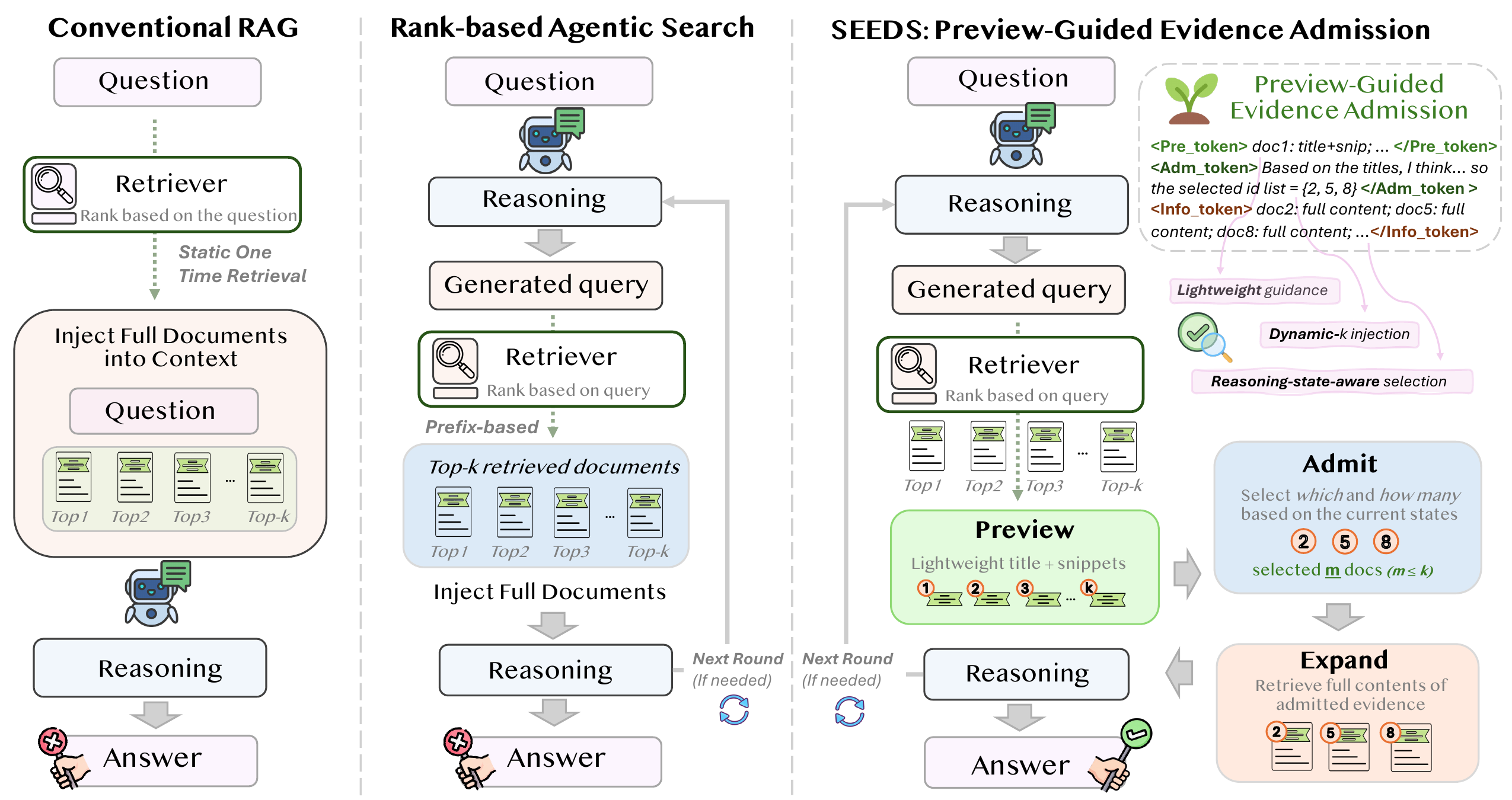}
    \caption{Overview of \seeds{} compared with conventional RAG and prefix rank-based agentic search. Conventional RAG injects top-ranked documents before reasoning, while agentic search retrieves iteratively but still injects a rank prefix at each round. \seeds{} presents lightweight previews, allowing the reasoning model to choose \emph{which candidates}, and \emph{how many}, to expand based on its current reasoning state. Only expanded documents enter subsequent reasoning rollouts.}
    \label{fig:main}
\end{figure}

This section presents \seeds{}, a preview-guided evidence admission framework for
multi-round search agents. We first formalize agentic search and the evidence
admission problem (\S\ref{sec:problem}). We then introduce the preview-guided
admission mechanism (\S\ref{sec:preview_admission}), which separates lightweight
candidate observation from detailed evidence consumption. Finally, we describe
the end-to-end inference procedure (\S\ref{sec:inference}).
Figure~\ref{fig:main} illustrates how \seeds{} inserts an explicit evidence
admission step between retrieval and full evidence consumption.

\subsection{Problem Formalization}
\label{sec:problem}

\paragraph{Agentic Search-Augmented Reasoning.}
We consider a multi-round search agent that solves an input question $q$ by
interleaving reasoning with external search. Given a task instruction $I$, the
model generates a reasoning trajectory
$\mathcal{R}=(\mathcal{R}_1,\ldots,\mathcal{R}_{T_{\mathcal R}})$
followed by a final answer
$a=(a_1,\ldots,a_{T_a})$, where $T_{\mathcal R}$ and $T_a$
denote the lengths of the reasoning trajectory and final answer, respectively.
During reasoning, the model may issue multiple search queries and incorporate
external evidence into its trajectory. Let $\mathcal{Z}_{<t}$ denote the
evidence incorporated before reasoning step $t$. The generation process can be formalized as a prediction process:
\begin{equation}
\begin{aligned}
P(\mathcal{R},a\mid I,q)
=&
\underbrace{
\prod_{t=1}^{T_{\mathcal R}}
P\!\left(
\mathcal{R}_t
\mid
\mathcal{R}_{<t},I,q,\mathcal{Z}_{<t}
\right)
}_{\text{Reasoning Process}}
&\cdot
\underbrace{
\prod_{t=1}^{T_a}
P\!\left(
a_t
\mid
a_{<t},\mathcal{R},I,q
\right)
}_{\text{Answer Generation}} .
\end{aligned}
\label{eq:reasoning_rollout}
\end{equation}
Evidence incorporated during search therefore becomes part of the evolving
reasoning state and can directly affect subsequent reasoning, future search
actions, and the final answer.

\paragraph{Retrieval and Evidence Admission.}
Suppose the model initiates its $i$-th search after accumulating reasoning
history $\mathcal{R}^{(<i)}$ and previously incorporated evidence
$\mathcal{Z}^{(<i)}$. It generates a search query $q_s^{(i)}$ and gets a ranked
candidate set $\mathcal{D}^{(i)}$ with size $K$ from the retriever $\operatorname{Ret(\cdot)}$:
\begin{equation}
q_s^{(i)}
\sim
P_\theta
\left(
\cdot
\mid
I,q,\mathcal{R}^{(<i)},\mathcal{Z}^{(<i)}
\right),
\qquad
\mathcal{D}^{(i)}
=
\operatorname{Ret}\!\left(q_s^{(i)}\right)
=
\left(
d_1^{(i)},\ldots,d_K^{(i)}
\right).
\label{eq:search_retrieve}
\end{equation}

Retrieval determines which evidence is \emph{available} to the agent, but not
which evidence should influence subsequent reasoning. We therefore introduce
an explicit \emph{evidence admission} decision. Let $\mathcal{H}^{(i)}$ denote
the agent's reasoning state before the admission decision, including its
reasoning history, previous search actions, and previously incorporated
evidence. Given the retrieved candidate set $\mathcal{D}^{(i)}$, the goal is to
select an admitted subset
$
\mathcal{E}^{(i)}
\subseteq
\mathcal{D}^{(i)}
$
whose detailed contents are expected to be useful for the agent's current
information need.

This formulation differs from fixed or adaptive rank-prefix strategies. A
fixed-$k$ strategy admits
$\{d_1^{(i)},\ldots,d_k^{(i)}\}$, while an adaptive-$k$ strategy changes the
prefix length. Evidence admission allows both the identity and the number of selected candidates to depend on the current reasoning state.

\subsection{Preview-Guided Evidence Admission}
\label{sec:preview_admission}

Directly deciding which candidates to admit presents a fundamental information
trade-off. The agent needs sufficient information to assess each candidate,
but reading every candidate in full would eliminate the benefit of selective
admission. \seeds{} addresses this problem by separating lightweight candidate
observation from detailed evidence consumption. It first exposes a compact
\emph{preview} of each retrieved candidate and then uses these previews together
with the current reasoning state to make the admission decision.

\paragraph{Candidate Preview.}
At search round $i$, each retrieved candidate
$d_j^{(i)} \in \mathcal{D}^{(i)}$, $j\in\{1, ..., K\}$ is represented by a lightweight preview:
\begin{equation}
\mathcal{P}^{(i)}
=
\left\{
p_1^{(i)},\ldots,p_K^{(i)}
\right\},
\qquad
p_j^{(i)}
=
\operatorname{Preview}
\left(
q_s^{(i)}, d_j^{(i)}
\right).
\label{eq:preview_set}
\end{equation}
The preview provides a low-cost observation of each candidate before its full
content is consumed. The preview operator depends only on the current search
query and retrieved document, and is independent of the agent's reasoning
trajectory. It summarizes information available from the
retrieval result itself, while the subsequent admission decision is conditioned
on the evolving reasoning state. 
For local corpus retrieval, we instantiate $\operatorname{Preview}(\cdot)$ as a deterministic, compact,
query-conditioned representation that preserves both document identity and locally
relevant content. Specifically, we concatenate the document title with a local text window selected by query-keyword matching, where the window with the highest query-keyword overlap is exposed as the preview:
\begin{equation}
p_j^{(i)}
=
\left[
\operatorname{Title}(d_j^{(i)});
\operatorname{Match}(q_s^{(i)}, d_j^{(i)})
\right].
\label{eq:preview_design}
\end{equation}
For live web search, we directly use the title and
snippet returned by the search engine as the candidate preview $\operatorname{Preview}(\cdot)$. Detailed
preview construction is provided in Appendix~\ref{app:snapshot}.

\paragraph{State-Conditioned Evidence Admission.}
Given the preview set $\mathcal{P}^{(i)}$, the agent decides which candidates
should be inspected in full content. We define the admission policy $\mathcal{A}$ as:
\begin{equation}
\mathcal{E}^{(i)}
=
\mathcal{A}
\left(
\mathcal{H}^{(i)},
\mathcal{P}^{(i)}
\right),
\label{eq:admission}
\end{equation}
Because
$\mathcal{A}$ conditions on the current reasoning state, both the composition
and cardinality of $\mathcal{E}^{(i)}$ can change as the agent acquires new
information.

Unlike rank-prefix selection, the admitted set need not contain the
highest-ranked candidates or follow any contiguous prefix of the retrieval
ranking. The action space therefore supports arbitrary subsets of
$\mathcal{D}^{(i)}$, including the empty set when no candidate warrants further
inspection. The admission policy can be instantiated directly by an existing
reasoning model through prompting, or further optimized through supervised or
reinforcement learning when training is available.

\paragraph{Evidence Expansion and Incorporation.}
Admission determines \emph{which} candidates should be consumed, while
expansion is the operation that reveals the full detailed contents of admitted candidates:
\begin{equation}
\mathcal{Z}^{(i)}
=
\operatorname{Expand}
\left(
\mathcal{E}^{(i)}
\right)
=
\left\{
\operatorname{Full}(d)
\mid
d\in\mathcal{E}^{(i)}
\right\}.
\label{eq:selected_expansion}
\end{equation}
Here, $\operatorname{Full}(d)$ denotes accessing the full content associated
with document $d$, which in our implementation corresponds to the retrieved passage or web page. $\operatorname{Expand}(\cdot)$ collects the full contents
of all admitted passages into the evidence set $\mathcal{Z}^{(i)}$.
This evidence is then incorporated into the reasoning
trajectory and becomes part of the state used for subsequent reasoning and
search, corresponding to the evidence term in the
reasoning process of Eq.~\ref{eq:reasoning_rollout}.

Overall, \seeds{} decomposes evidence acquisition within agentic search into the following steps:
$
\mathcal{D}^{(i)}
\;\xrightarrow{\operatorname{Preview}}\;
\mathcal{P}^{(i)}
\;\xrightarrow[\mathcal{H}^{(i)}]{\mathcal{A}}\;
\mathcal{E}^{(i)}
\;\xrightarrow{\operatorname{Expand}}\;
\mathcal{Z}^{(i)}.
\label{eq:seeds_pipeline}
$ This decomposition separates evidence availability, lightweight observation, admission, and detailed document consumption.

\subsection{Inference}
\label{sec:inference}

At inference time, the search agent iteratively generates a query and retrieves a candidate evidence pool at each search round. As shown in Algorithm~\ref{alg:seeds}, the agent first observes lightweight previews of the retrieved candidates and then selects an admitted subset conditioned on its current reasoning state. Only the admitted candidates are expanded in full and incorporated into the reasoning trajectory. This process repeats until the model produces a final answer.

\vspace{-2em}
\begin{algorithm}[H]
\captionsetup{font=small}
\algsetup{linenosize=\scriptsize}
\caption{\seeds{} inference}
\label{alg:seeds}
\scriptsize
\begin{algorithmic}[1]
\REQUIRE Reasoning model $\mathcal{M}$, question $q$, retriever $\operatorname{Ret}$, operators $\operatorname{Preview}$ and $\operatorname{Expand}$

\STATE Initialize reasoning history $\mathcal{H}$ with $q$

\WHILE{true}
    \STATE Generate reasoning and a search or answer action $u$ using $\mathcal{M}(\mathcal{H})$
    \STATE Append the generated reasoning and $u$ to $\mathcal{H}$

    \IF{$u$ is a final answer}
        \STATE \textbf{break}
    \ENDIF

    \STATE Extract the search query $q_s^{(i)}$ from $u$
    \STATE Retrieve candidates
    $\mathcal{D}^{(i)} \leftarrow \operatorname{Ret}(q_s^{(i)})$

    \STATE Construct previews
    $\mathcal{P}^{(i)} \leftarrow
    \operatorname{Preview}(q_s^{(i)}, \mathcal{D}^{(i)})$

    \STATE Append $\mathcal{P}^{(i)}$ to $\mathcal{H}$

    \STATE Select admitted evidence
    $\mathcal{E}^{(i)} \leftarrow
    \mathcal{M}(\mathcal{H})$,
    where $\mathcal{E}^{(i)} \subseteq \mathcal{D}^{(i)}$

    \STATE Expand admitted evidence
    $\mathcal{Z}^{(i)} \leftarrow
    \operatorname{Expand}(\mathcal{E}^{(i)})$

    \STATE Append $\mathcal{Z}^{(i)}$ to $\mathcal{H}$
\ENDWHILE
\RETURN $\mathcal H$

\end{algorithmic}
\end{algorithm}
\vspace{-2em}

\begingroup
\setlength{\textfloatsep}{6pt plus 2pt minus 2pt}
\setlength{\floatsep}{6pt plus 2pt minus 2pt}
\setlength{\intextsep}{4pt plus 2pt minus 2pt}
\captionsetup{font=small,skip=3pt}
\section{Experiments}
\label{sec:exp}

We evaluate \seeds{} through four research questions (RQs): \textbf{RQ1:}
Can \seeds{} improve the effectiveness of multi-round search agents while reducing full-evidence consumption across diverse reasoning benchmarks and search environments?
\textbf{RQ2:}
Does preview-guided evidence admission select more useful evidence than prefix rank-based strategies?
\textbf{RQ3:}
How robust is \seeds{} to noisy or distracting retrieval results, and how sensitive is it to the preview design?
\textbf{RQ4:}
Does \seeds{} generalize across reasoning backbones and agentic search paradigms, and can its evidence admission policy be further improved through learning?


\subsection{Experiment Setup}

\paragraph{Datasets.}
We evaluate \seeds{} on nine benchmarks spanning two settings.
\textbf{\emph{(i) Open-domain QA tasks}}, including \emph{single-hop} benchmarks NQ~\citep{nqbench} and TriviaQA~\citep{triviaqa}, and \emph{multi-hop} benchmarks  HotpotQA~\citep{yang2018hotpotqa}, 2Wiki~\citep{2wiki}, MuSiQue~\citep{musique}, and Bamboogle~\citep{press2023bamb}. \textbf{\emph{(ii) Challenging search-intensive tasks}}, including BrowseComp~\citep{wei2025browsecomp}, BrowseComp-zh~\citep{zhou2025browsecompzh}, and Humanity's Last Exam (HLE)~\citep{2026hle}.
More details are provided in Appendix~\ref{app:datasets}.

\paragraph{Baselines.}
We compare against four baseline families.
\textbf{\emph{(i) No retrieval}} uses Direct Reasoning without external evidence.
\textbf{\emph{(ii) Single-round RAG}} includes fixed-$k$ RAG~\citep{lewis2021rag} and the adaptive retrieval method MAIN-RAG~\citep{2025mainrag}.
\textbf{\emph{(iii) Multi-round agentic search}} includes ReAct~\citep{yao2023react} with fixed $k$ and a dynamic-$k$ variant (\emph{dyn-$k$}), where the reasoning model determines how many top-ranked documents to consume at each search round without previews. We also extend MAIN-RAG to the multi-round setting, denoted MAIN-RAG (multi-round).
\textbf{\emph{(iv) Agentic search with document refinement}} uses Search-o1~\citep{li2025searcho1}, representing a different agentic search paradigm that performs additional reasoning over retrieved documents. 

\paragraph{Implementation.}
Our main experiments use QwQ-32B~\citep{qwq32b} as the main reasoning backbone model for evaluation, with $K{=}10$ retrieved candidates per search.
For open-domain QA under \textbf{\emph{local corpus retrieval}}, we use the 2018 Wikipedia dump~\citep{localwikidump} as the knowledge corpus and E5~\citep{wang2024e5} for retrieval, and report Exact Match (EM).
For the challenging benchmarks under \textbf{\emph{live web search}}, we use the Serper API for search and the Jina Reader API to fetch web-page contents, with LLM-judged accuracy reported. We also report token-level F1 on the predicted answers, and the average number of fully accessed documents or web pages per question. We instantiate our preview-guided evidence admission mechanism within two different agentic search paradigms, ReAct and Search-o1, yielding \textbf{\seeds{}-ReAct} and \textbf{\seeds{}-o1}, respectively. In the local corpus retrieval setting, the preview length is about 60 tokens to cover the title and query-aware matched text. Detailed prompts and implementation are provided in Appendices~\ref{app:setup} and \ref{app:prompts}.

\subsection{Main Results (RQ1)}
\label{sec:rq1}

\begin{table}[!htbp]
\caption{Main results (EM\%, F1\%, \#Doc per question) of \seeds{} compared with baselines on open-domain QA benchmarks using the local Wikipedia dump corpus, with the reasoning backbone QwQ-32B model.}
\label{tab:main_results}
\centering\small\setlength{\tabcolsep}{3pt}
\resizebox{\linewidth}{!}{%
\begin{tabular}{lcccccccccccc|ccc}
\toprule
& \multicolumn{4}{c}{Single-hop QA} & \multicolumn{8}{c}{Multi-hop QA} & \multicolumn{3}{c}{Avg.}\\
\cmidrule(lr){2-5}\cmidrule(lr){6-13}\cmidrule(lr){14-16}
Method & \multicolumn{2}{c}{NQ} & \multicolumn{2}{c}{TriviaQA} & \multicolumn{2}{c}{HotpotQA} & \multicolumn{2}{c}{2Wiki} & \multicolumn{2}{c}{MuSiQue} & \multicolumn{2}{c}{Bamboogle} & \multicolumn{3}{c}{(all)}\\
\cmidrule(lr){2-3}\cmidrule(lr){4-5}\cmidrule(lr){6-7}\cmidrule(lr){8-9}\cmidrule(lr){10-11}\cmidrule(lr){12-13}\cmidrule(lr){14-16}
& EM & F1 & EM & F1 & EM & F1 & EM & F1 & EM & F1 & EM & F1 & EM$\uparrow$ & F1$\uparrow$  & \#Doc$\downarrow$ \\
\midrule
\multicolumn{16}{l}{\emph{Without Retrieval}}\\
Direct Reasoning & 28.0 & 39.7 & 66.4 & 72.0 & 27.0 & 36.6 & 32.6 & \textbf{38.5} & 10.2 & 19.2 & 48.0 & 63.0 & 35.4 & 44.8 & 0\\
\midrule
\multicolumn{16}{l}{\emph{Retrieval-augmented Reasoning}}\\
RAG $k=3$ & 39.8 & 49.7 & 60.8 & 66.6 & 31.0 & 39.7 & 22.0 & 28.9 & 8.4 & 17.4 & 32.8 & 45.0 & 32.5 & 41.2 & 3.0\\
RAG $k=10$ & 42.8 & \textbf{53.2} & 64.2 & 70.8 & 30.4 & 39.9 & 25.8 & 31.9 & 10.0 & 19.1 & 28.8 & 40.5 & 33.7 & 42.6 & 10.0\\
MAIN-RAG & \textbf{43.0} & 53.0 & 64.0 & 70.3  & 31.6 & 40.2  &  24.2 & 31.2  &  14.0 & 23.0  & 32.8 & 44.9  & 34.9 & 43.8 & 9.8\\
\midrule
\multicolumn{16}{l}{\emph{Multi-round Agentic Search}}\\
ReAct $k=3$ & 35.2 & 45.2 & 64.2 & 72.2 & 32.0 & 44.7 & 27.4 & 34.3 & 13.2 & 21.8 & 52.8 & 63.9 & 37.5 & 47.0 & 7.7\\
ReAct $k=10$ & 34.8 & 45.1 & 63.8 & 71.8 & 28.8 & 39.9 & 21.0 & 26.5 & 11.4 & 19.4 & 52.0 & 63.6 & 35.3 & 44.4 & 14.9\\
ReAct dyn-$k$ & 33.0 & 43.1 & 65.0 & 73.3 & 31.6 & 43.8 & 25.6 & 35.4 & 14.8 & 25.6 & 51.2 & 65.6 & 36.9 & 47.8 & 7.7\\
MAIN-RAG (multi-round) & 35.6 & 45.1 & 65.4 & 73.3  & 33.0  & 45.7 & 28.2 & 36.2 & 14.0  & 23.0 & 56.0 & 68.0  & 38.7 & 48.6 & 11.0 \\
\rowcolor{seedsbg!50}\textbf{\seeds{}-ReAct \emph{(ours)}} & 39.6 & 51.7 & \textbf{69.6} & \textbf{76.9} & \textbf{35.6} & \textbf{47.2} & \textbf{31.2} & 38.1 & \textbf{16.6} & \textbf{28.0} & \textbf{56.8} & \textbf{68.6} & \textbf{41.6} & \textbf{51.8} & \textbf{3.1}\\
\midrule
\multicolumn{16}{l}{\emph{Multi-round Agentic Search with Reason-in-Documents}}\\
Search-o1 & \textbf{40.6} & \textbf{53.7} & 68.2 & \textbf{76.8} & 38.4 & 49.8 & 32.2 & 39.4 & \textbf{16.6} & 27.4 & 59.2 & 71.0 & 42.5 & 53.0 & 30.2\\
\rowcolor{seedsbg!50} \textbf{\seeds{}-o1 \emph{(ours)}} & 40.2 & 52.9 & \textbf{68.8} & 76.4 & \textbf{39.8} & \textbf{51.4} & \textbf{33.6} & \textbf{42.0} & 16.4 & \textbf{27.9} & \textbf{60.8} & \textbf{71.4} & \textbf{43.3} & \textbf{53.7} & \textbf{12.6}\\
\bottomrule
\end{tabular}}
\end{table}

Tables~\ref{tab:main_results} and~\ref{tab:web_results} report results under local-corpus and live-web search. Here are main observations:

\paragraph{\seeds{} improves answer accuracy while reducing full-document consumption.}
As shown in Table~\ref{tab:main_results}, \seeds{}-ReAct achieves the highest EM on all six local Wikipedia benchmarks among the multi-round agentic search methods without Reason-in-Documents. Its average EM exceeds single-round MAIN-RAG by \textbf{+6.7} points and the multi-round baselines by \textbf{+2.9$\sim$6.3} points. Full-document consumption decreases by approximately \textbf{$\downarrow$60--79\%} relative to the multi-round baselines. When integrated with Search-o1, \seeds{}-o1 slightly increases average EM while reducing full-document consumption from 30.2 to 12.6 documents per question (approximately \textbf{$\downarrow$58\%}). These results show that selective evidence admission improves answer quality while consuming fewer full documents.

\begin{wraptable}{r}{0.65\linewidth}
\centering
\vspace{-0.5em}
\caption{
Results (Acc\%, F1\%, \#Pages per question) on web browsing and expert-level question benchmarks with live web search, QwQ-32B.
}
\label{tab:web_results}
\scriptsize
\setlength{\tabcolsep}{2.4pt}
\renewcommand{\arraystretch}{0.92}
\resizebox{\linewidth}{!}{
\begin{tabular}{lcccccc|ccc}
\toprule
& \multicolumn{2}{c}{BrowseComp} 
& \multicolumn{2}{c}{BrowseComp-zh} 
& \multicolumn{2}{c}{HLE} 
& \multicolumn{3}{c}{Avg.} \\
\cmidrule(lr){2-3}
\cmidrule(lr){4-5}
\cmidrule(lr){6-7}
\cmidrule(lr){8-10}
Method 
& Acc & F1 
& Acc & F1 
& Acc & F1 
& Acc$\uparrow$ & F1$\uparrow$ & \#Pages$\downarrow$ \\
\midrule
Search-o1 
& 3.32 & 6.09 
& 14.19 & 21.30 
& 10.34 & \textbf{10.60} 
& 9.28 & 12.66 & 27.81 \\
\textbf{\seeds{}-o1 \emph{(ours)}} 
& \textbf{5.29} & \textbf{8.42} 
& \textbf{16.26} & \textbf{23.85} 
& \textbf{13.79} & 6.66 
& \textbf{11.78} & \textbf{12.98} & \textbf{8.30} \\
\bottomrule
\end{tabular}
}
\end{wraptable}

\paragraph{The benefits extend to live-web search.}
Results in Table~\ref{tab:web_results} show that \seeds{} remains effective on more challenging live-web benchmarks with more complex questions, where retrieved candidates come from open-web search and are less controlled in quality and relevance. Nevertheless, \seeds{}-o1 improves average accuracy by \textbf{+2.5} points while reducing the number of fetched full web pages by approximately \textbf{$\downarrow$70\%}. These results indicate that preview-guided evidence admission generalizes beyond local retrieval and remains effective in live-web search.

\subsection{Evidence Admission Behavior and Efficiency (RQ2)}
\label{sec:rq2}

A central question for \seeds{} is whether preview-guided admission improves the \emph{quality and composition} of evidence entering the reasoning trajectory, rather than merely adapting how much evidence is consumed. We evaluate this through admission behavior, evidence quality, and consumption efficiency, using the same retrieval pool size $K{=}10$ for both prefix-based baselines and \seeds{} variants 
\begin{wrapfigure}{r}{0.43\columnwidth}
    \centering
    \includegraphics[width=\linewidth]{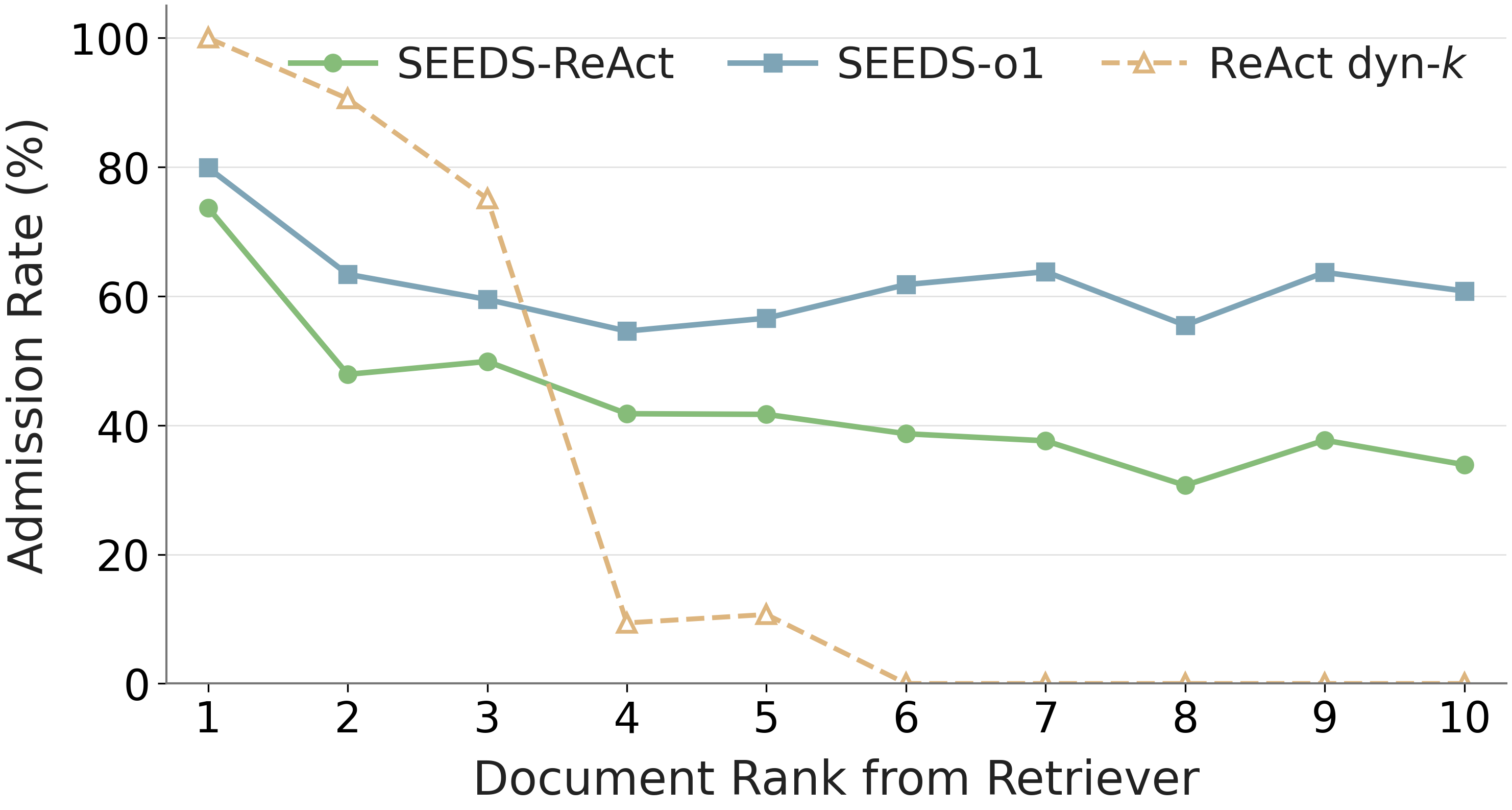}
    \caption{Evidence admission probability (\%) of documents with different retrieval ranks.}
    \label{fig:rank_selection}
    \vspace{-2em}
\end{wrapfigure}
with local corpus retrieval settings. To assess admission quality, we mark a document as \textbf{\emph{answer-bearing}} if its normalized full text contains any normalized gold-answer alias. 
Detailed answer-bearing matching rules are provided in Appendix~\ref{app:expansion_quality}.

\paragraph{Non-prefix Evidence Admission.}
We first test whether \seeds{} changes \emph{which} retrieved candidates are admitted, rather than only admitting those top-ranked documents. Across more than 5k search rounds of data collection on the open-domain QA benchmarks, Fig.~\ref{fig:rank_selection} shows that \seeds{}-ReAct and \seeds{}-o1 admit evidence throughout the whole retrieved candidate pool, including those lower-ranked documents. In contrast, ReAct dyn-$k$ selects almost exclusively from the highest-ranked candidates (rank $<$ 4), with admission probability dropping sharply beyond the top few ranks, with almost 0\% admission probability for documents ranked starting from 6. As summarized in Table~\ref{tab:rankdist}, $44.1\%$ and $39.6\%$ of documents admitted by \seeds{}-ReAct and \seeds{}-o1 fall beyond their corresponding evidence set size $|S|$ (top-$|S|$ prefixes), and their selected sets depart from these prefixes in roughly $80\%$ of search rounds. Thus, \seeds{} adapts not only admission cardinality, but also the flexible composition of the evidence entering the reasoning trajectory.

\begin{figure}[!htbp]
\centering
\begin{subfigure}[t]{0.46\linewidth}
    \centering
    \includegraphics[width=\linewidth]{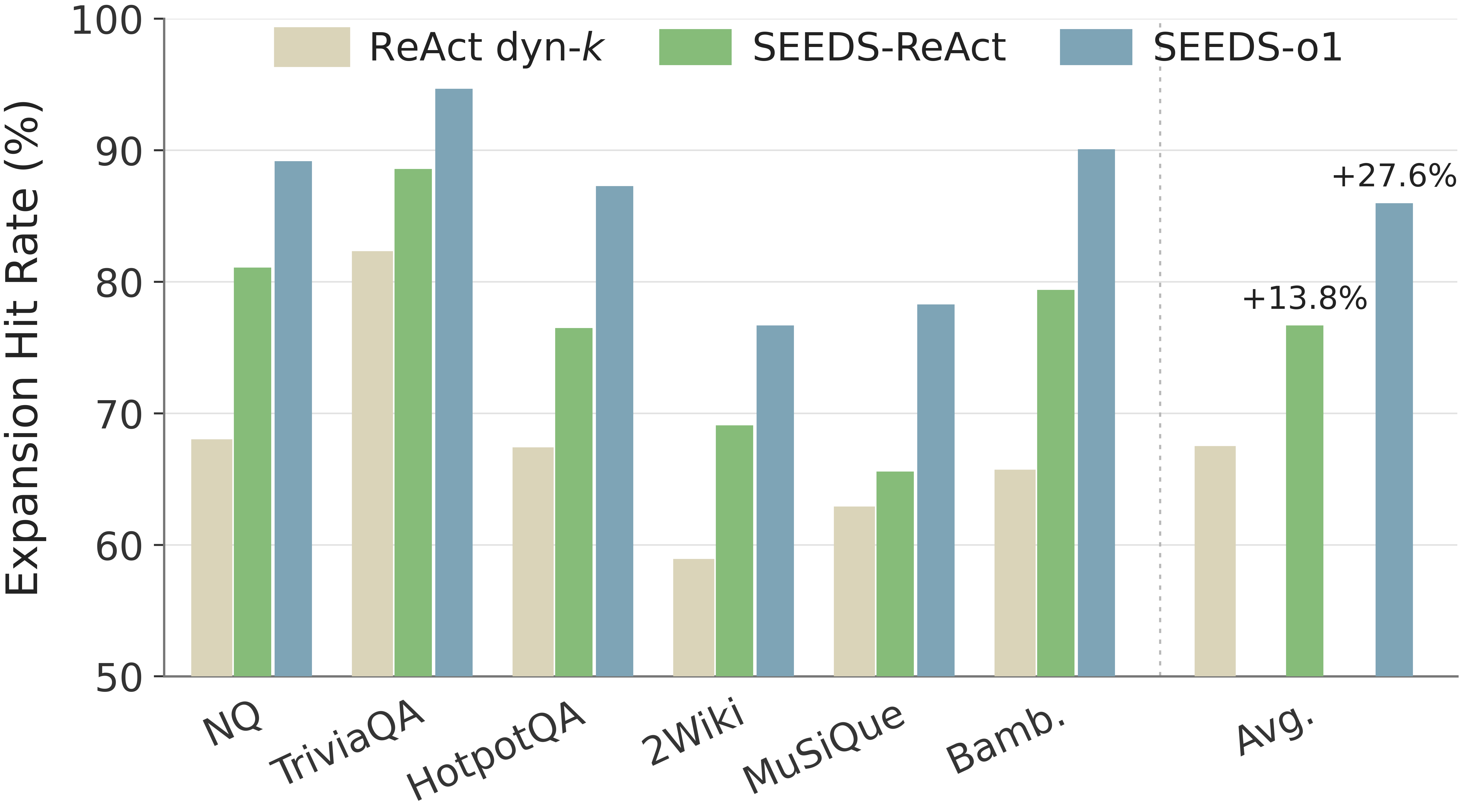}
    \caption{Answer-bearing evidence hit rate.}
    \label{fig:exp_quality}
\end{subfigure}
\hfill
\begin{subfigure}[t]{0.46\linewidth}
    \centering
    \includegraphics[width=\linewidth]{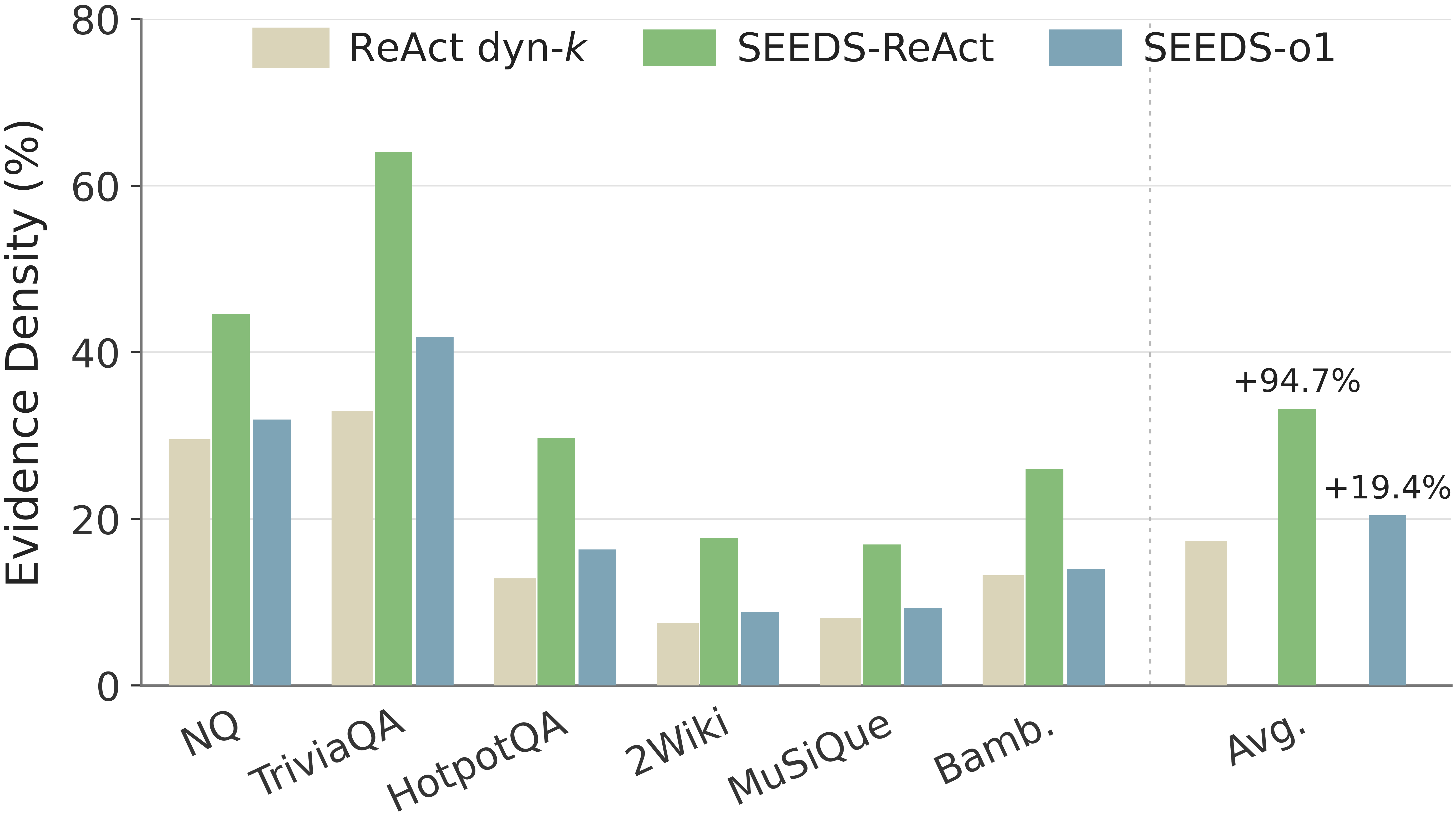}
    \caption{Answer-bearing evidence density.}
    \label{fig:exp_density}
\end{subfigure}

\caption{
Evidence admission quality.
(a) \emph{Hit rate} is the probability that an answer-bearing document is admitted when it is available in the preview candidate pool.
(b) \emph{Density} is the proportion of answer-bearing documents among all documents admitted into the reasoning trajectory.
}
\label{fig:expansion_quality}
\end{figure}

\paragraph{Evidence Admission Quality.}
We next examine whether this greater flexibility leads to more useful evidence being admitted. When answer-bearing evidence is available in the candidate pool, \seeds{} admits it more reliably than the rank-prefix baseline across all six open-domain QA benchmarks. As shown in Fig.~\ref{fig:expansion_quality}(a), \seeds{}-ReAct and \seeds{}-o1 achieve average relative improvements of $13.8\%$ and $27.6\%$ in answer-bearing evidence hit rate over ReAct dyn-$k$, respectively, highlighting the benefit of preview guidance for evidence selection. Moreover, Fig.~\ref{fig:expansion_quality}(b) shows that the admitted sets contain a substantially higher density of answer-bearing evidence, indicating that \seeds{} preserves more useful information while filtering unnecessary content. Together, these results show that preview-guided admission improves both the likelihood of admitting useful evidence and the quality of the evidence that ultimately enters the reasoning trajectory.

\paragraph{Evidence Consumption Efficiency.}
We examine whether selective admission also reduces evidence-acquisition cost. As shown in Table~\ref{tab:efficiency}, \seeds{}-ReAct reduces both search frequency and full-document consumption, while cutting document-token usage by $58\%$ relative to ReAct dyn-$k$. Similarly, \seeds{}-o1 uses nearly the same number of searches as Search-o1 but reduces full-document reads and document-token consumption by about $58\%$. These results show that \seeds{} improves efficiency by concentrating full-document access on fewer, more useful evidence.

\vspace{0.3em}
\begin{table}[!htbp]
\centering
\begin{minipage}[t]{0.4\linewidth}
\vspace{0pt}
\centering
\captionof{table}{Admitted answer-bearing evidence with rank beyond the selected evidence size.}
\label{tab:rankdist}
\small
\resizebox{\linewidth}{!}{%
\begin{tabular}{lcc}
\toprule
Method & Documents & Rounds \\
\midrule
ReAct dyn-$k$ & 0.0\% & 0.0\% \\
\midrule
\textbf{\seeds{}-ReAct} & 44.1\% & 77.7\% \\
\textbf{\seeds{}-o1}    & 39.6\% & 80.2\% \\
\bottomrule
\end{tabular}%
}
\end{minipage}\hfill%
\begin{minipage}[t]{0.55\linewidth}
\vspace{0pt}
\centering
\captionof{table}{Evidence consumption. The number of searches, documents, and document tokens per question are reported.}
\label{tab:efficiency}
\scriptsize
\setlength{\tabcolsep}{3.2pt}
\resizebox{\linewidth}{!}{%
\begin{tabular}{lcccc}
\toprule
Method
& Searches $\downarrow$
& Full docs $\downarrow$
& Docs/search $\downarrow$
& Doc.\ tokens $\downarrow$ \\
\midrule
ReAct dyn-$k$ & 2.67 & 7.7 & 2.93 & 1267 \\
\textbf{\seeds{}-ReAct}
& \textbf{2.00} & \textbf{3.1} & \textbf{1.78} & \textbf{528} \\
\midrule
Search-o1
& \textbf{3.02} & 30.2 & 10.00 & 5006 \\
\textbf{\seeds{}-o1}
& 3.03 & \textbf{12.6} & \textbf{4.25} & \textbf{2075} \\
\bottomrule
\end{tabular}%
}
\end{minipage}
\end{table}

\subsection{Robustness and Sensitivity (RQ3)}
\label{sec:robustness}

\paragraph{Robustness to Retrieval Perturbations.}
We evaluate whether preview-guided admission can filter distracting evidence before it enters the reasoning trajectory. For each top-$K$ candidate pool, we replace a fraction $\rho$ of \emph{non-answer-bearing} documents with same-topic distractors sampled from outside the top-$K$, while always retaining the available answer-bearing evidence. Increasing $\rho$ therefore creates noisier candidate pools without removing the evidence required to answer the question.

\begin{table}[!htbp]
\centering
\vspace{-1em}
\caption{
Robustness comparison (EM \%) under preview candidate-pool perturbations. 
}
\label{tab:robustness}
\small
\setlength{\tabcolsep}{10pt}
\begin{tabular}{lcccc}
\toprule
Method & Clean & $\rho=30\%$ & $\rho=50\%$ & $\rho=80\%$ \\
\midrule
ReAct dyn-$k$
& 36.9 & 37.0 (+0.1) & 36.4 (-0.5) & 33.4 (-3.5) \\
ReAct $k=3$
& 37.5 & 38.2 (+0.7) & 36.0 (-1.5) & 36.8 (-0.7) \\
ReAct $k=10$
& 35.3 & 36.0 (+0.7) & 35.7 (+0.4) & 36.9 (+1.6) \\
\midrule
\textbf{\seeds{}-ReAct}
& \textbf{41.6}
& \textbf{42.1} (+0.5)
& \textbf{42.3} (+0.7)
& \textbf{42.9} (+1.3) \\
\quad Injected noise admitted
& & 6.4\% & 7.5\% & 7.3\% \\
\bottomrule
\end{tabular}
\end{table}

As shown in Table~\ref{tab:robustness}, rank-prefix methods with smaller evidence sets are more sensitive to retrieval perturbations. ReAct dyn-$k$ and ReAct $k{=}3$ degrade as more high-ranked candidates are replaced because they cannot selectively bypass distractors within the consumed prefix. ReAct $k{=}10$ is more stable, likely because the larger context preserves sufficient useful evidence despite the perturbation. In contrast, \seeds{}-ReAct admits fewer than $8\%$ of injected distractors across all settings while maintaining or improving accuracy. At $\rho=0.8$, it improves by $1.3$ EM over clean retrieval, suggesting that preview-guided admission can still isolate useful evidence under heavy perturbation.

\paragraph{Sensitivity to Preview Candidate-Pool Size.}
We study how the preview pool size affects answer quality and search behavior by varying $N$ from $5$ to $20$ based on \seeds{}-ReAct. As shown in Fig.~\ref{fig:k_pool_sensitivity}, accuracy remains relatively stable, indicating limited sensitivity to $N$. Larger pools lead the agent to admit more documents per question while generally requiring fewer search rounds, suggesting that more evidence can be incorporated within each round. The effect on accuracy is task-dependent: single-hop NQ slightly benefits from larger pools, whereas multi-hop HotpotQA degrades at $N=20$, likely due to additional distracting evidence. We further conduct experiments on various preview designs and include analyses in Appendix~\ref{app:snapshot}.

\begin{figure}[!htbp]
\centering
\begin{minipage}[t]{0.6\linewidth}
    \centering
    \includegraphics[width=1\linewidth]{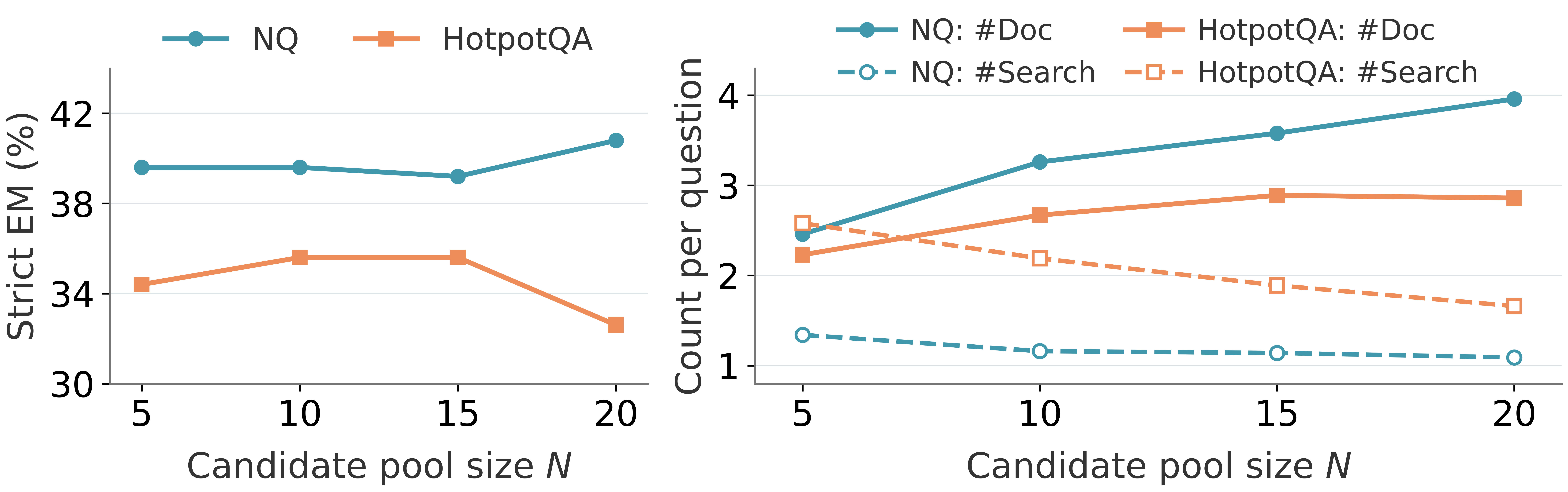}
    \caption{
    Impact of the size of the preview candidate pool.
    }
    \label{fig:k_pool_sensitivity}
\end{minipage}
\hfill
\begin{minipage}[t]{0.38\linewidth}
  
    \centering
    \includegraphics[width=1\linewidth]{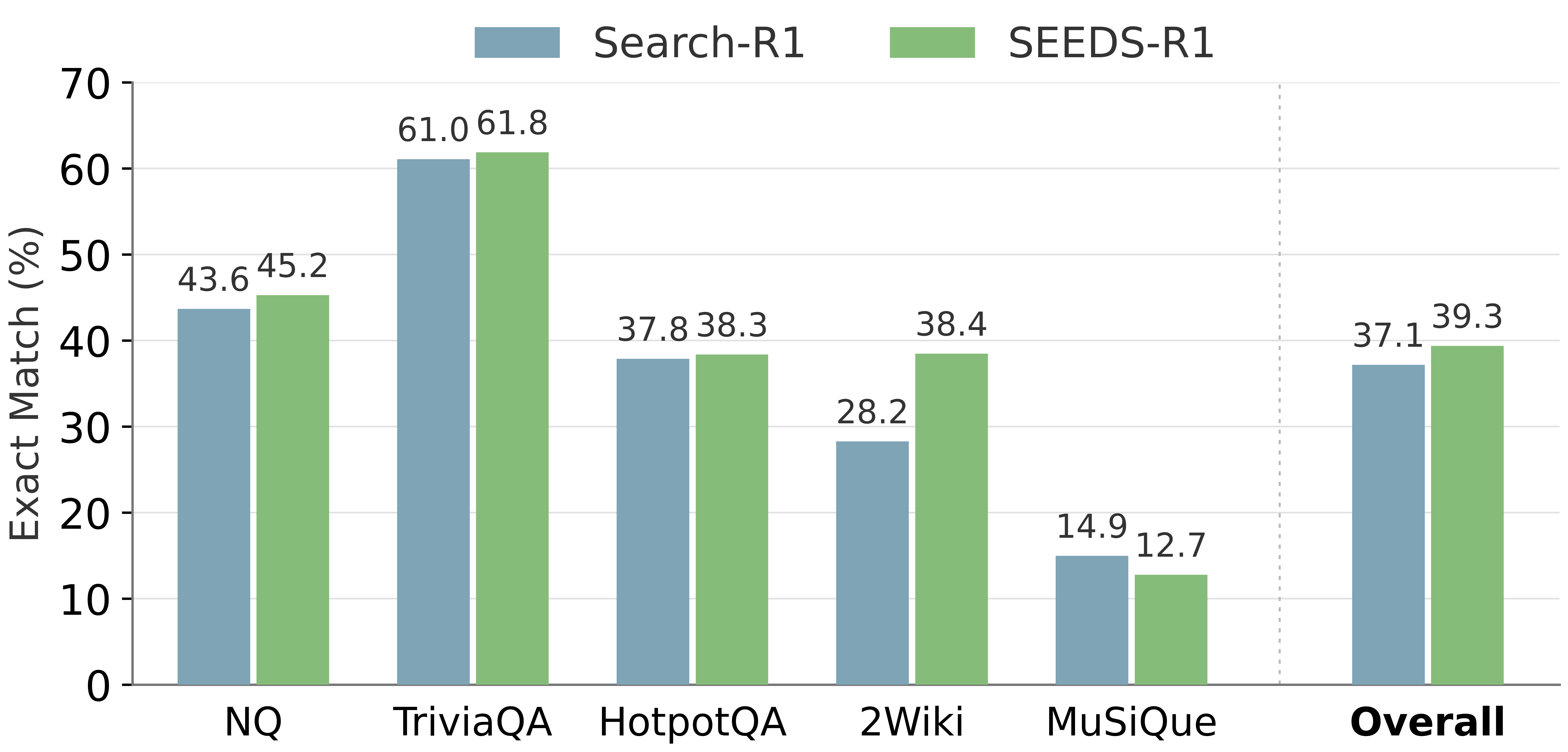}
    \caption{The learned variant \seeds{}-R1.}
    \label{fig:rl_analysis}
\end{minipage}
\end{figure}

\subsection{Generalization and Learning (RQ4)}
\label{sec:rq4}

\paragraph{Generalization across Reasoning Backbones.}
Table~\ref{tab:backbone} evaluates \seeds{} with DeepSeek-V4.1-Flash~\citep{deepseekai2026deepseekv41flash}, which improves accuracy over QwQ-32B on all four benchmarks. Compared with Search-o1 under the same DeepSeek backbone, \seeds{} gains 1.0 EM point on HotpotQA and 1.2 on 2Wiki, matches HLE accuracy, and loses 1.4 EM points on NQ. It reduces full-document access by $52$--$61\%$ on local QA and fetched pages by approximately $44\%$ on HLE. These results show that \seeds{} benefits from stronger reasoning models while preserving its evidence-efficiency advantage.

\begin{table}[!htbp]
\centering
\caption{
Generalization across different reasoning backbones and search scenarios.
}
\label{tab:backbone}
\small
\setlength{\tabcolsep}{5pt}
\begin{tabular}{l*{4}{cc}}
\toprule
& \multicolumn{2}{c}{NQ}
& \multicolumn{2}{c}{HotpotQA}
& \multicolumn{2}{c}{2Wiki}
& \multicolumn{2}{c}{HLE} \\
\cmidrule(lr){2-3}
\cmidrule(lr){4-5}
\cmidrule(lr){6-7}
\cmidrule(lr){8-9}
Configuration
& $\uparrow$EM & $\downarrow$\#Doc
& $\uparrow$EM & $\downarrow$\#Doc
& $\uparrow$EM & $\downarrow$\#Doc
& $\uparrow$Acc. & $\downarrow$\#Pages \\
\midrule
\seeds{}-o1 (QwQ-32B)
& 40.2 & 9.8
& 39.8 & 13.2
& 33.6 & 16.6
& 13.8 & \textbf{8.3} \\

\seeds{}-o1 (DeepSeek)
& 43.7 & \textbf{9.2}
& \textbf{45.2} & \textbf{12.2}
& \textbf{58.4} & \textbf{14.6}
& \textbf{45.1} & 12.9 \\

Search-o1 (DeepSeek)
& \textbf{45.1} & 19.2
& 44.2 & 26.2
& 57.2 & 37.6
& \textbf{45.1} & 22.9 \\
\bottomrule
\end{tabular}
\end{table}

\paragraph{Extension to a Learnable Setting.}
Finally, we explore whether evidence admission can also be optimized through
training. We
train a Qwen2.5-7B model with the \seeds{}-ReAct interaction on NQ and
HotpotQA, using SFT followed by PPO with both final-answer and
evidence-selection rewards, yielding \seeds{}-R1. As shown in
Fig.~\ref{fig:rl_analysis}, \seeds{}-R1 improves average EM from $37.1$ to
$39.3$, with particularly strong gains on 2Wiki, although improvements are
mixed across datasets. These results provide initial evidence that preview-guided
admission can be incorporated into learned search policies, while more effective
admission-specific training remains an open direction. Additional training
details are provided in Appendix~\ref{app:training}.

\par\endgroup

\section{Conclusion}
\label{sec:conclusion}

This work studies an overlooked decision in agentic search: which retrieved evidence should enter and influence the evolving reasoning trajectory. We formulate this problem as \emph{evidence admission} and introduce \seeds{}, a preview-guided framework for reasoning-aware selection before full evidence consumption. Across nine knowledge-intensive QA benchmarks under both local-corpus and live-web search, \seeds{} improves search effectiveness while substantially reducing full-evidence consumption. Further analyses show that \seeds{} admits useful evidence beyond retrieval prefixes, more reliably recovers answer-bearing evidence, and remains effective across retrieval perturbations and reasoning backbones. These findings highlight evidence admission as a complementary dimension of agentic search and suggest more effective admission policies.

\bibliography{main}
\bibliographystyle{iclr2027_conference}

\appendix

\clearpage
\begingroup
\hypersetup{hidelinks}
\raggedbottom
\setlength{\textfloatsep}{10pt plus 2pt minus 2pt}
\setlength{\floatsep}{8pt plus 2pt minus 2pt}
\setlength{\intextsep}{10pt plus 2pt minus 2pt}
\renewcommand{\topfraction}{0.95}
\renewcommand{\bottomfraction}{0.85}
\renewcommand{\textfraction}{0.05}
\renewcommand{\floatpagefraction}{0.8}
\setcounter{topnumber}{5}
\setcounter{bottomnumber}{3}
\setcounter{totalnumber}{6}

\section{Experimental Setup and Interaction Protocol}
\label{app:setup}
\label{app}

\subsection{Datasets}
\label{app:datasets}

The main evaluation covers six local-corpus QA benchmarks and three live-web
benchmarks. NQ~\citep{nqbench} and
TriviaQA~\citep{triviaqa} evaluate factual question answering, while
HotpotQA~\citep{yang2018hotpotqa}, 2WikiMultiHopQA
(2Wiki)~\citep{2wiki}, MuSiQue~\citep{musique}, and
Bamboogle~\citep{press2023bamb} require combining information across
multiple reasoning steps. We use all 125 data points of Bamboogle dataset~\citep{press2023bamb} and randomly select 500 data points from other datasets. Averages over these six datasets are macro-averages unless
otherwise stated.

BrowseComp~\citep{wei2025browsecomp}, BrowseComp-zh~\citep{zhou2025browsecompzh},
and Humanity's Last Exam (HLE)~\citep{2026hle} evaluate more demanding
search-intensive questions. The live-web evaluation contains 1266, 289, and
435 questions, respectively. The HLE evaluation is restricted to a randomly chosen subset containing 435 questions out of 2158 text-only
questions. We report judged answer accuracy using Qwen2.5-72B-Instruct~\citep{qwen2025qwen25} as the evaluator, with the subject-level HLE
breakdown in Appendix~\ref{app:web}.

\subsection{Baselines}
\label{app:baselines}

The baseline families are defined in Section~\ref{sec:exp}. Fixed-$k$ RAG
retrieves once, whereas ReAct repeats retrieval with $k\in\{3,10\}$ or a
model-selected prefix length. MAIN-RAG filters candidates once or within the
multi-round loop. Search-o1 refines full documents through Reason-in-Documents.
\seeds{}-ReAct adds admission before direct reading and \seeds{}-o1 adds before
refinement. Following Search-o1~\citep{li2025searcho1}, we adopt a back-off strategy for both Search-o1 and \seeds{}-o1, where the corresponding direct-reasoning output is used for evaluation when the agent fails to produce a final answer. 

\subsection{Retrieval and Inference Settings}
\label{app:implementation}

\paragraph{Local corpus.}
The main experiments use QwQ-32B~\citep{qwq32b} and the Wikipedia-2018 corpus
used by Search-R1, containing approximately 21 million passages of 100 words
each. We encode passages with e5-base-v2~\citep{wang2024e5} and retrieve by exact
FAISS inner-product search. Each query returns $K=10$ candidates. Selected
passages are exposed up to a limit of 1,500 characters per passage~\citep{jin2025searchr1}.

\paragraph{Live web.}
We use the Serper API for search and the Jina Reader API for page retrieval,
as in the main experiments. Search results provide titles and snippets for the
preview pool. For each selected result, we retain up to 3,000 characters of
page content centered around the matched snippet. Page-fetch counts record
cache misses, rather than repeated accesses to cached pages.

\paragraph{Additional backbone.}
The DeepSeek experiments use the \texttt{deepseek-flash} endpoint at
\url{https://api.deepseek.com/v1}, with \texttt{reasoning\_effort="high"}
and thinking enabled through
\texttt{extra\_body=\{"thinking":\{"type":"enabled"\}\}}.
The trained Qwen2.5-7B setting has a separate interaction budget and is described
in Appendix~\ref{app:training}.

\subsection{Prompt Instructions and Interaction Templates}
\label{app:interaction_templates}
In the implementation, we set the special preview token as <snap>, and each candidate's identifier, title, and matched snippet are appended in
\texttt{<snap>...</snap>}. Selection identifiers refer to this most recent
pool. Only selected passages are subsequently returned as full evidence, and 
the model never supplies its own retrieved observations.

\paragraph{Admission instructions.}
The template below summarizes the protocol, rather than reproducing a
particular run verbatim. \seeds{}-ReAct stores selections as
\texttt{selected\_ids: [i, j, ...]} and returns passages in
\texttt{<obs>...</obs>}; the trained environment instead uses
\texttt{<information>...</information>}.

\begin{quote}\small
Issue a query in \texttt{<search>...</search>} and stop for the previews.
After each pool, emit one \texttt{<expand>...</expand>} action naming the
passages whose full text could help answer the question. Compare candidates
with the evidence already collected. Do not reject a relevant passage merely
because its preview omits the answer. Skip irrelevant or redundant passages;
selecting none is allowed. Stop for the selected evidence, then search again
or give a concise answer in \texttt{<answer>...</answer>}.
\end{quote}

\paragraph{Baseline interfaces.}
Fixed-$k$ ReAct issues \texttt{<search>[query]</search>} and receives the top-$k$
full passages in \texttt{<information>...</information>} without a preview or
expansion decision. ReAct dyn-$k$ instead emits
\texttt{<search>topk=N, query='[query]'</search>}, with $1\leq N\leq10$;
its chosen documents remain a retrieval-rank prefix. Thus, the dynamic-$k$ baseline adapts reading depth without observing previews
or selecting outside the retrieval prefix.

\section{Preview Construction and Ablation}
\label{app:snapshot}

\paragraph{Query-aware previews.}
For local retrieval, a preview combines the passage title with a short
query-aware view of its text. The first 80 characters cover the title, and a 240-character window exposes a span with high lexical overlap with
the current search query. The total preview is capped at 60 tokens. The construction
uses the query and passage text, requires no additional model call, and does
not use gold answers or the agent's reasoning history. For live-web retrieval,
the search engine's title and snippet serve directly as the preview.

\paragraph{Representation and Length.}
For local corpus retrieval, each retrieved ``document'' corresponds to a
pre-segmented Wikipedia passage (chunk). Table~\ref{tab:snap_design}
compares several preview representations, including the title only, the first
$n$ tokens of the chunk, a query-aware local window, and a one-line chunk
summary. We use Qwen2.5-32B-Instruct~\citep{qwen2025qwen25} for summarization. No single representation consistently performs best across datasets,
and increasing preview length does not monotonically improve accuracy.
We adopt the
query-aware window because it provides compact, query-relevant information
without requiring an additional summarization model.

\begin{table}[H]
\centering
\caption{Preview ablation: strict EM (\%, QwQ-32B, $n=500$ per dataset).}
\label{tab:snap_design}
\small
\setlength{\tabcolsep}{7pt}
\begin{tabular}{@{}lrrrr@{}}
\toprule
Preview & NQ & TriviaQA & HotpotQA & 2Wiki \\
\midrule
Title only & 38.0 & 67.2 & 33.8 & 37.8 \\
Prefix, 30 tokens & 37.4 & 67.2 & 35.2 & \textbf{39.2} \\
Prefix, 60 tokens & 38.6 & 67.0 & \textbf{37.6} & 32.4 \\
Prefix, 90 tokens & \textbf{41.0} & 68.0 & 35.6 & 28.6 \\
Query window (adopted) & 39.6 & \textbf{69.6} & 35.6 & 31.2 \\
Chunk summary & 38.2 & 68.6 & 35.8 & 35.2 \\
\bottomrule
\end{tabular}
\end{table}

\section{Evidence Admission Analysis and Efficiency}
\label{app:evidence}

\subsection{Answer-Bearing Evidence Proxy}
\label{app:expansion_quality}

We use answer occurrence as a document-level proxy. 
A document is \emph{answer-bearing} if its normalized full text contains any
normalized gold-answer alias. Normalization lowercases text, removes diacritics
and punctuation, and collapses whitespace. 
This proxy measures answer occurrence, not evidential sufficiency. It can miss
bridge passages that do not contain the final answer and can count unrelated
mentions of an answer string. The answer-bearing hit rate in
Figure~\ref{fig:expansion_quality}(a) is therefore interpreted conditionally on such a
document being available in the candidate pool. For the non-prefix analysis, a selected document lies beyond the matched
prefix when its retrieval rank exceeds the number of documents admitted in
that round. This comparison measures departure from rank-based admission at
the observed reading budget.

\subsection{Per-Dataset Evidence Consumption}
\label{app:efficiency}

Tables~\ref{tab:app_search_calls}--\ref{tab:app_document_tokens} decompose the
aggregate costs in Table~\ref{tab:efficiency}. Search calls count retrieval
invocations; full-document reads count passages passed to the reasoning model
or the Reason-in-Documents reader. Document-token counts use the model
tokenizer and exclude prompts, previews, reasoning, and generated output.
They quantify full-evidence consumption, rather than total inference tokens
or latency. Each average is the macro-average across the six local benchmarks.

\seeds{}-ReAct inspects about three documents per question across datasets,
while fixed-$k$ agents read more as search frequency increases. For
\seeds{}-o1, the search frequency is close to Search-o1, but document reads
and document-token consumption are lower on every dataset. The per-dataset
breakdown thus supports the main-text distinction between searching less
and reading less per search.

\begin{table}[H]
\caption{
Average number of search calls per question on QwQ-32B.
Avg.\ is the macro-average across the six benchmarks.
}
\label{tab:app_search_calls}
\centering
\small
\setlength{\tabcolsep}{5pt}
\begin{tabular}{@{}lrrrrrrr@{}}
\toprule
Method
& NQ
& TriviaQA
& HotpotQA
& 2Wiki
& MuSiQue
& Bamboogle
& Avg. \\
\midrule
ReAct $k=3$
& 1.70 & 1.32 & 2.75 & 4.47 & 3.02 & 2.06 & 2.55 \\
ReAct $k=10$
& 0.96 & 0.89 & 1.63 & 2.66 & 1.70 & 1.10 & 1.49 \\
ReAct dyn-$k$
& 1.77 & 1.53 & 2.71
& 5.33 & 2.74 & 1.92
& 2.67 \\
\seeds{}-ReAct
& 1.16 & 1.23 & 2.19
& 3.26 & 2.41 & 1.78
& 2.00 \\
\midrule
Search-o1
& 2.04 & 1.84 & 3.37
& 4.73 & 3.60 & 2.55
& 3.02 \\
\seeds{}-o1
& 2.07 & 1.80 & 3.38
& 4.52 & 3.72 & 2.70
& 3.03 \\
\bottomrule
\end{tabular}
\end{table}

\begin{table}[H]
\caption{
Average number of full documents inspected per question.
A document is counted when its full content is passed to a reader.
Avg.\ is the macro-average across the six benchmarks.
}
\label{tab:app_full_documents}
\centering
\small
\setlength{\tabcolsep}{5pt}
\begin{tabular}{@{}lrrrrrrr@{}}
\toprule
Method
& NQ
& TriviaQA
& HotpotQA
& 2Wiki
& MuSiQue
& Bamboogle
& Avg. \\
\midrule
RAG $k=3$
& 3.0 & 3.0 & 3.0 & 3.0 & 3.0 & 3.0 & 3.0 \\
RAG $k=10$
& 10.0 & 10.0 & 10.0 & 10.0 & 10.0 & 10.0 & 10.0 \\
ReAct $k=3$
& 5.1 & 4.0 & 8.3 & 13.4 & 9.1 & 6.2 & 7.7 \\
ReAct $k=10$
& 9.6 & 8.9 & 16.3 & 26.6 & 17.0 & 11.0 & 14.9 \\
ReAct dyn-$k$
& 5.7 & 4.7 & 7.9
& 14.3 & 8.0 & 5.3
& 7.7 \\
\seeds{}-ReAct
& 3.3 & 3.3 & 2.7
& 2.9 & 3.0 & 3.2
& 3.1 \\
\midrule
Search-o1
& 20.4 & 18.4 & 33.7
& 47.3 & 36.0 & 25.5
& 30.2 \\
\seeds{}-o1
& 9.8 & 8.1 & 13.2
& 16.6 & 15.8 & 12.0
& 12.6 \\
\bottomrule
\end{tabular}
\end{table}

\begin{table}[H]
\caption{
Average number of full-document tokens consumed per question,
measured with the model tokenizer.
The metric excludes reasoning, preview, query, and generated-output
tokens. Avg.\ is the macro-average across six benchmarks.
}
\label{tab:app_document_tokens}
\centering
\small
\setlength{\tabcolsep}{5pt}
\begin{tabular}{@{}lrrrrrrr@{}}
\toprule
Method
& NQ
& TriviaQA
& HotpotQA
& 2Wiki
& MuSiQue
& Bamboogle
& Avg. \\
\midrule
RAG $k=3$
& 452 & 454 & 475 & 502 & 467 & 460 & 468 \\
RAG $k=10$
& 1,512 & 1,518 & 1,584 & 1,671 & 1,563 & 1,526 & 1,562 \\
ReAct $k=3$
& 806 & 622 & 1,370 & 2,270 & 1,482 & 972 & 1,254 \\
ReAct $k=10$
& 1,518 & 1,405 & 2,704 & 4,438 & 2,767 & 1,744 & 2,429 \\
ReAct dyn-$k$
& 902 & 747 & 1,336
& 2,448 & 1,308 & 861
& 1,267 \\
\seeds{}-ReAct
& 512 & 513 & 461
& 527 & 511 & 643
& 528 \\
\midrule
Search-o1
& 3,310 & 2,959 & 5,634
& 8,095 & 5,911 & 4,125
& 5,006 \\
\seeds{}-o1
& 1,568 & 1,280 & 2,185
& 2,931 & 2,580 & 1,909
& 2,075 \\
\bottomrule
\end{tabular}
\end{table}

\section{Detailed Live-Web Results}
\label{app:web}

Table~\ref{tab:hle} breaks down the 435-question text-only HLE evaluation
by subject. \seeds{}-o1 answers 60 questions correctly, compared with 45 for
Search-o1, with higher accuracy in seven subjects and a tie in Engineering.
The small subject subsets should be interpreted descriptively.

\begin{table}[H]
\centering
\caption{HLE text-only accuracy (\%) by subject (QwQ-32B). $n$ is the number of
questions.}
\label{tab:hle}
\small
\setlength{\tabcolsep}{12pt}
\begin{tabular}{@{}lrrr@{}}
\toprule
Subject & $n$ & Search-o1 & \seeds{}-o1 \\
\midrule
Mathematics & 187 & 12.83 & \textbf{13.90} \\
Computer science & 52 & 9.62 & \textbf{13.46} \\
Humanities & 45 & 13.33 & \textbf{17.78} \\
Biology / medicine & 42 & 11.90 & \textbf{19.05} \\
Other & 42 & 7.14 & \textbf{11.90} \\
Physics & 40 & 0.00 & \textbf{5.00} \\
Chemistry & 14 & 14.29 & \textbf{28.57} \\
Engineering & 13 & 0.00 & 0.00 \\
\midrule
Overall & 435 & 10.34 & \textbf{13.79} \\
\bottomrule
\end{tabular}
\end{table}

\section{Training and Evaluation of \seeds{}-R1}
\label{app:training}
\label{sec:training}

\seeds{}-R1 uses Qwen2.5-7B-Base and is trained in two stages: supervised
fine-tuning (SFT) teaches the search--preview--expand--answer protocol, and
proximal policy optimization (PPO)~\citep{schulman2017ppo} then optimizes answer correctness and
evidence selection. Table~\ref{tab:train-hparams} summarizes the configuration.

\subsection{Interaction Environment}
\label{sec:train-env}

The training environment uses the same Wikipedia corpus and E5 retriever as
the main experiments, with $K=5$ candidates per query. It returns titles and
query-aware previews in \texttt{<snap>} blocks and exposes only selected
passages in \texttt{<information>} blocks. Each search must be followed by
one expand decision, selecting between zero and five passages. Episodes are
limited to four retrieval rounds and ten policy actions; observations are
capped at 1,200 tokens per step. The final response is enclosed in
\texttt{<answer>...</answer>}.

\subsection{Stage 1: Supervised Warm Start}
\label{sec:train-sft}

\paragraph{Demonstration construction.}
We collect QwQ-32B trajectories on NQ and HotpotQA training questions using
the same five-candidate environment. The teacher uses temperature 0.6,
top-$p$ 0.95, top-$k$ 5, and up to 4,096 generated tokens per turn.
Teacher \texttt{<think>} blocks are removed before storage, retaining actions,
observations, and the final answer.

A trajectory is accepted if it satisfies the protocol and either produces an
exact-match answer or exposes a gold answer in the observations shown to the
teacher. The latter condition uses whole-word matching after SQuAD-style
normalization. The final answer is then replaced by the gold string, while
the teacher's original answer is retained for auditing. This is an
answer-occurrence filter, not a guarantee of evidential sufficiency.
We collect 1,000 accepted trajectories, and the split contains 900 training trajectories
(420 NQ, 480 HotpotQA) and 100 development trajectories (47 NQ, 53 HotpotQA).

\paragraph{Objective and checkpoint selection.}
We fine-tune all student parameters with causal language-modeling loss on
policy-generated action and answer tokens only:
\begin{equation}
\mathcal{L}_{\mathrm{SFT}}
=-\mathbb{E}_{\tau^*\sim\mathcal{D}_{\mathrm{SFT}}}
\left[\sum_{j\in\mathcal{M}}
\log p_\theta(y_j^*\mid y_{<j}^*,x)\right],
\label{eq:sft}
\end{equation}
where $x$ is the question and instruction, and $\mathcal{M}$ indexes policy
tokens. Instructions and environment observations are masked out of the loss.
Development EM increases from 0.14 before SFT to 0.49 after one epoch and
0.53 after five epochs (0.55 on NQ, 0.51 on HotpotQA), with protocol compliance
at least 0.96. Although development loss is lowest after epoch 1, we select
epoch 5 by development EM and use it to initialize both the PPO actor and critic.

\subsection{Stage 2: Reinforcement Learning}
\label{sec:train-rl}

\paragraph{Evidence labels.}
The outcome reward is final-answer exact match,
$R_{\mathrm{EM}}(\tau)\in\{0,1\}$. An auxiliary selection bonus evaluates
which documents the agent opens. For NQ, a document is labeled positive if it
contains a normalized gold answer; unexpanded candidates are judged only from
their visible previews. For HotpotQA, positive documents have titles matching
annotated supporting-fact titles, so bridge documents receive credit even when
they omit the final answer. Other HotpotQA documents containing the answer are
neutral. Positive documents are deduplicated by normalized title across the
trajectory. These labels are available during training only.

\paragraph{Selection reward.}
Let $S_r$ be the documents expanded at round $r$, $\mathcal{G}$ and
$\mathcal{N}$ the positive and neutral documents, and $\mathcal{O}_{<r}$ the
positive documents opened before that round. Let $\mathcal{V}$ be positive
documents visible in previews and $\mathcal{O}$ those opened anywhere in the
trajectory. The selection reward is
\begin{align}
\Delta_r
&=\alpha\bigl|(S_r\cap\mathcal{G})\setminus\mathcal{O}_{<r}\bigr|
-\beta\bigl|S_r\setminus(\mathcal{G}\cup\mathcal{N})\bigr|,
\label{eq:sel-event}\\
\Delta_{\mathrm{miss}}
&=-\gamma\bigl|\mathcal{V}\setminus\mathcal{O}\bigr|
\mathbb{1}[R_{\mathrm{EM}}(\tau)=0],
\label{eq:sel-miss}\\
R_{\mathrm{sel}}(\tau)
&=\operatorname{clip}\!\left(\sum_r\Delta_r+\Delta_{\mathrm{miss}},-c,c\right),
\label{eq:sel-total}
\end{align}
with $(\alpha,\beta,\gamma,c)=(0.15,0.05,0.15,0.6)$.
The reward credits the first opening of each positive document and penalizes
opening negative documents. Missed visible evidence is penalized only when
the final answer is incorrect, allowing the policy to answer from a preview
without an unnecessary expansion. Clipping bounds the auxiliary contribution.

Each selection term is assigned to the last token of its expand action;
the missed-evidence term is assigned to the last expand action, or the final
generated token if no expansion occurs. The combined selection contribution
is clipped as above. Exact match is assigned to the final generated token.
A per-token KL penalty with coefficient 0.01 regularizes the actor toward the
frozen SFT reference policy. The PPO objective uses these token-level rewards
with generalized advantage estimation ($\gamma_{\mathrm{GAE}}=\lambda=1$).

\paragraph{Observation integrity and optimization.}
Reward computation parses only environment-inserted spans identified by the
observation mask, so policy-generated text cannot masquerade as retrieved
evidence. If parsed information blocks and accepted expansion actions disagree,
the selection bonus is discarded while the EM reward is retained. Environment
tokens are masked out of both policy and value losses. Evaluation uses EM
without the selection bonus.

PPO uses the NQ and HotpotQA training sets (79,168 and 90,447 questions,
respectively). The critic is updated alone for the first ten steps.
Training runs for 500 updates on two H200 GPUs, and
the reported model is the step-450 checkpoint. Optimization settings are
listed in Table~\ref{tab:train-hparams}.

\begin{table}[!htbp]
\centering
\small
\caption{Optimization settings for \seeds{}-R1 (Qwen2.5-7B). The interaction environment is specified in Appendix~\ref{sec:train-env}.}
\label{tab:train-hparams}

\setlength{\tabcolsep}{3pt}
\begin{tabular}{@{}lll@{}}
\toprule
 & Hyperparameter & Value \\
\midrule
\multirow{7}{*}{Stage 1: SFT}
 & Teacher / student & QwQ-32B / Qwen2.5-7B-Base \\
 & Training / dev trajectories & 900 / 100 \\
 & Loss & masked NLL on policy tokens only \\
 & Optimizer & AdamW, $\beta=(0.9,0.95)$, weight decay 0.01 \\
 & Learning rate / warmup & $1\times10^{-5}$ / 10\% \\
 & Batch size / max.\ length & 16 / 8{,}192 tokens \\
 & Epochs & 5 (selected by dev EM) \\
\midrule
\multirow{11}{*}{Stage 2: RL}
 & Algorithm & PPO + GAE ($\gamma=\lambda=1$) \\
 & Training questions & NQ + HotpotQA (169{,}615) \\
 & Batch / mini-batch & 256 / 128 \\
 & Actor LR / warmup & $5\times10^{-7}$ / 2\%, constant \\
 & Critic LR / warmup & $1\times10^{-5}$ / 1.5\%, constant \\
 & Critic-only warmup & 10 updates \\
 & PPO clip / value clip / entropy & 0.2 / 0.5 / 0.001 \\
 & KL penalty ($\beta_{\mathrm{KL}}$, in reward) & 0.01 \\
 & Selection bonus $(\alpha,\beta,\gamma,c)$ & (0.15, 0.05, 0.15, 0.6) \\
 & Rollout temperature / tokens per turn & 1.0 / 500 \\
 & Updates & 500 (checkpoint at 450 used) \\
\bottomrule
\end{tabular}
\end{table}

\subsection{Training Process}
\label{app:training_results}

Figure~\ref{fig:app_training_curves} shows the selection bonus and answer
exact match during PPO training.
During critic warm-up, the SFT policy opens 32\% of visible positive documents
on NQ and 42\% on HotpotQA, with selection precision 0.32 and 0.35.
Over updates 91--110, these open rates rise to 74\% and 81\%, while document
reads increase from 1.30 to 2.34 per question. Over updates 426--450, open
rates reach 74\% and 88\%, precision reaches 0.43 and 0.46, and reads decrease
to 2.12 documents per question. These window averages indicate an initial
increase in evidence acquisition followed by more selective reading.

\begin{figure}[!htbp]
\centering
\begin{subfigure}[t]{0.48\linewidth}
\centering
\includegraphics[width=\linewidth]{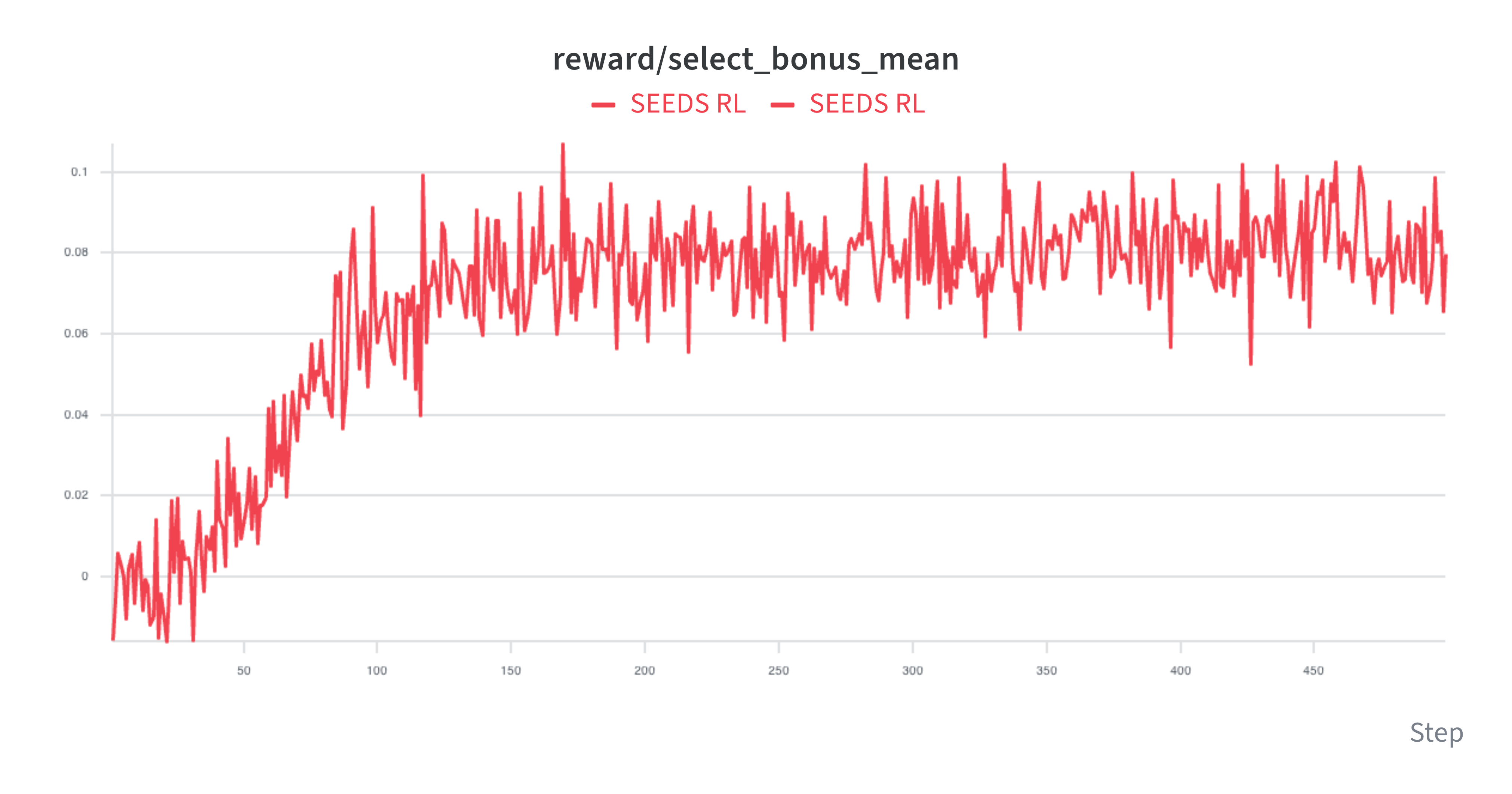}
\caption{Mean selection bonus.}
\label{fig:app_selection_bonus}
\end{subfigure}
\hfill
\begin{subfigure}[t]{0.48\linewidth}
\centering
\includegraphics[width=\linewidth]{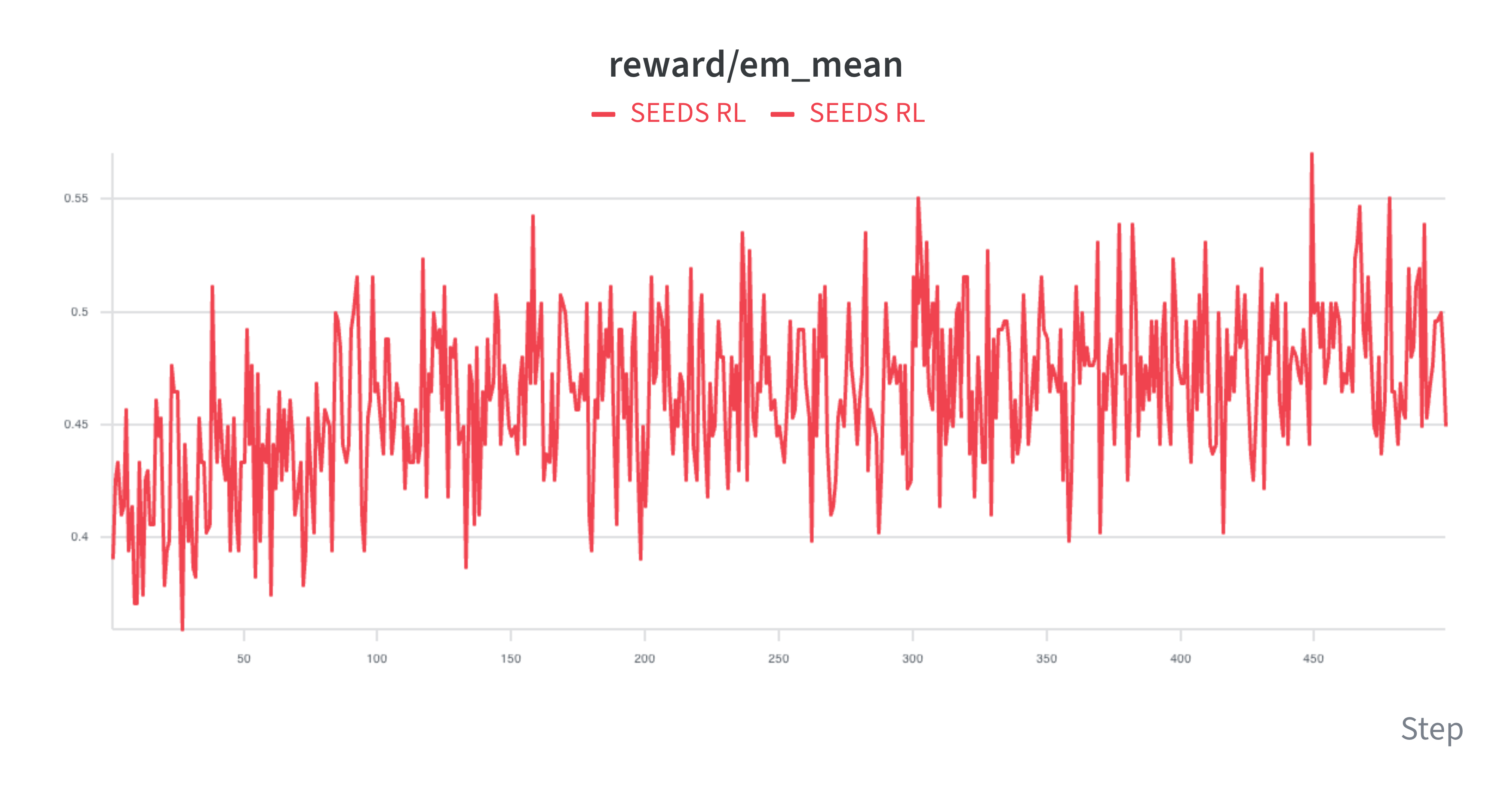}
\caption{Mean answer exact match.}
\label{fig:app_training_em}
\end{subfigure}
\caption{Training curves for \seeds{}-R1 (Qwen2.5-7B, $K=5$).
The horizontal axis is the optimizer update; the first ten updates train only
the critic. Curves show training-batch means, not held-out evaluation scores.}
\label{fig:app_training_curves}
\end{figure}

\section{Prompt Templates}
\label{app:prompts}

\subsection{SEEDS-o1}
\label{app:prompt-seeds-o1}

SEEDS-o1 adds a preview-guided evidence admission module to the Search-o1 agent prompt.
Specifically, in our implementation, after each search query, the environment returns a \texttt{<snap>} pool of
\verb|{N}| candidates, each shown as an id, a title, and a short snippet.
The agent must then emit exactly one \texttt{<expand>} block that selects up to
\verb|{K}| of them. Only the selected passages go to the Search-o1
Reason-in-Documents module, whose output is returned in
\texttt{<|begin\_search\_result|>} \ldots\ \texttt{<|end\_search\_result|>}.
\verb|{T}| caps the number of search calls. The expand instruction is the
\emph{max-information} variant: the number of expanded passages is not fixed,
but the agent is told to be inclusive when selecting.

{\scriptsize
\begin{verbatim}
You are a reasoning assistant with the ability to perform web searches to help you answer
  the user's question accurately. You have special tools:

- To perform a search: write <|begin_search_query|> your query here <|end_search_query|>.
Then, the system will return a snapshot pool of {N} candidate passages in the format <snap>
  ...snapshot... </snap>. Each entry has an id, a title, and a short snippet (not the full
  content).

- After receiving the snapshot, decide which passages are worth reading in full: write
  <expand> a rationale for your selection, then a final line selected_ids: [i, j, ...]
  </expand>.
Use only ids from the most recent <snap>. Select as many passages as carry any useful
  information for the question -- be inclusive and prefer expanding a passage over skipping
  it, so you gather as much useful information as you can (between 0 and {K}). Only omit a
  passage if it is clearly irrelevant or fully redundant with ones you already chose. You
  still MUST output <expand>...</expand> after every <snap>. Do not answer directly after
  <snap>, and do not issue another search before <expand>.
Then, the system will analyze the selected passages and provide you with helpful information
  in the format <|begin_search_result|> ...search results... <|end_search_result|>.

You can repeat the search process multiple times if necessary. The maximum number of search
  attempts is limited to {T}.

Once you have all the information you need, continue your reasoning.

Example:
Question: "Alice David is the voice of Lara Croft in a video game developed by which
  company?"
Assistant thinking steps:
- I need to find out who voices Lara Croft in the video game.
- Then, I need to determine which company developed that video game.

Assistant:
<|begin_search_query|>Alice David Lara Croft voice<|end_search_query|>

(System returns a snapshot pool of candidate passages inside <snap>...</snap>)

Assistant:
<expand>Snippet [2] looks most relevant to who voices Lara Croft. selected_ids: [2]</expand>

(System returns processed information from the selected passages)

Assistant thinks: The search results indicate that Alice David is the voice of Lara Croft in
  a specific video game. Now, I need to find out which company developed that game.

Assistant:
<|begin_search_query|>video game developed by Alice David Lara Croft<|end_search_query|>

(System returns a snapshot pool of candidate passages inside <snap>...</snap>)

Assistant:
<expand>Snippet [1] names the developer of the game. selected_ids: [1]</expand>

(System returns processed information from the selected passages)

Assistant continues reasoning with the new information...

Remember:
- Use <|begin_search_query|> to request a web search and end with <|end_search_query|>.
- After each search, the system replies with a <snap> snapshot pool; you MUST immediately
  use <expand>...</expand> ending with a `selected_ids: [...]` line before answering or
  searching again.
- When done searching, continue your reasoning.
- When you have enough information, write the final answer inside <answer>...</answer>. Keep
  it SHORT -- a few words or a single phrase (a name, date, or entity), NOT a full sentence.

Question: {question}

\end{verbatim}
}

\subsection{SEEDS-ReAct}
\label{app:prompt-seeds-react}

{\scriptsize
\begin{verbatim}
You are a reasoning assistant that answers questions with the help of a search tool.

You follow this four-tag protocol; the system parses your output and injects results between turns:

1. Reason step by step inside `<think>...</think>` whenever it helps.
2. When you need external evidence, issue one query inside `<search>...</search>`.
   After you close `</search>`, stop generating; the system will respond with a snapshot pool of 
   10 candidate passages inside `<snap>...</snap>`. Each entry has an id and a short snippet.
3. After receiving `<snap>`, choose which passages to read in full. Write `<expand>...</expand>` 
containing:
   - first, your reasoning over the snapshot — which passages could help answer the question and 
   why, then
   - a final line `selected_ids: [i, j, ...]` listing the ids to expand (a list of integers from 
   the most recent `<snap>`).
   Each snippet shows only the beginning of its passage, so the evidence you need may be in the 
   part you cannot see. Select as many ids as carry any useful information for the question — 
   be inclusive and prefer expanding a passage over skipping it, so you gather as much useful 
   information as you can (between 0 and 10). Only omit a passage if it is clearly irrelevant 
   or fully redundant with ones you already chose.
   After you close `</expand>`, stop generating; the system will respond with the full passage 
   content of the selected ids inside `<obs>...</obs>` (empty if you selected none).
4. Once you have enough evidence, write the final answer inside `<answer>...</answer>` and stop.

Rules:
- After each `<search>`, the system returns one `<snap>`; write exactly one `<expand>` after 
each `<snap>`.
- Keep the final `<answer>` short (a few words or a single phrase).
- If a passage's title names the very entity or work your current query asks about, expand it — 
even if its snippet does not show the fact you need.
- Passages open with names, titles and dates; facts such as place of death, spouse, or country 
are usually in the middle of the passage. Never skip a passage merely because its snippet does 
not contain the answer.
- IMPORTANT: think as long as you need, but every turn MUST end with exactly one protocol tag —
`<search>...</search>`, `<expand>...</expand>`, or `<answer>...</answer>`. Describing a planned 
search in your reasoning is not a substitute for emitting the `<search>` tag itself.

Example
-------
Question: When was the actor who played James Bond in Skyfall born?

<think>I need to (a) find who played James Bond in Skyfall, then (b) look up ...</think>
<search>actor who played James Bond in Skyfall</search>
<snap>
[1] Skyfall — Skyfall is the 23rd James Bond film, starring Daniel Craig as Bond...
[2] James Bond (character) — James Bond is a fictional British Secret Service agent...
[3] Daniel Craig — Daniel Wroughton Craig is an English actor known for ...
[4] Sean Connery — Thomas Sean Connery was a Scottish actor ...
...
[10] No Time to Die — No Time to Die is a 2021 spy film ...
</snap>
<expand>
Snippet [1] confirms Daniel Craig played Bond in Skyfall, and [3] is Craig's bio ... 
Snippet [4] carries a birth date but for the wrong actor (Connery), clearly irrelevant...
selected_ids: [3, 1]
</expand>
<obs>
[3] Daniel Craig — Daniel Wroughton Craig was born on 2 March 1968 ...
[1] Skyfall — Skyfall is the 23rd James Bond film...
</obs>
<think>[1] confirms the Skyfall Bond actor is ...</think>
<answer>2 March 1968</answer>

Now answer the following question.
Question: {question}
\end{verbatim}
}

\subsection{SEEDS-R1}
\label{app:prompt-seeds-r1}

SEEDS-R1 is trained and evaluated with the Search-R1~\citep{jin2025searchr1} instruction format, extended
with the \texttt{<snap>}/\texttt{<expand>} actions. The first paragraph below is
the prompt stored in the training data. The block of required-expand rules is
inserted before \texttt{Question:} when the data are loaded, instantiated with
the training limits (at most 4 search rounds and at most 5 documents per
expand). The same prompt is used for supervised warm-up, RL, and evaluation,
and is wrapped in the Qwen2.5 chat template
(system message: \texttt{You are a helpful assistant.}).

{\scriptsize
\begin{verbatim}
Answer the given question. You must conduct reasoning inside <think> and </think> first
  every time you get new information. After reasoning, if you find you lack some knowledge,
  you can call a search engine by <search> query </search>. The system will return the top-k
  document snapshots between <snap> and </snap>,
  where each snapshot is prefixed with an id. After reading the snapshots, you must reason
  and select which documents to read in full by writing <expand> reasoning: your reasoning |
  expand_ids: [id1, id2, ...] </expand> (or expand_ids: [] if no snapshot is worth
  expanding). The full content of the selected documents will be returned between
  <information> and </information>. You can search and expand as many times as you want. If
  you find no further external knowledge needed, you can directly provide the answer inside
  <answer> and </answer> without detailed illustrations. For example, <answer> xxx
  </answer>.

SEEDS REQUIRED-EXPAND RULES FOR THIS RUN (these override any earlier general wording):
- You may perform at most 4 search rounds.
- After every <snap> result, your next valid action MUST be exactly one <expand> reasoning:
  ... | expand_ids: [...] </expand>.
- You MUST NOT issue another <search> or an <answer> until that expand action has completed.
  The environment will not insert or force the action. Completing it earns the next
  configured search-expand pair bonus; skipping it earns no expand bonus for that search
  round.
- Use expand_ids: [] when no snapshot is worth opening; this empty choice still completes
  the required expand step.
- Select at most 5 document ids in one expand.
- A direct answer is allowed before any search. After a search, an answer is allowed only
  after its required expand.

Question: {question}
\end{verbatim}
}

\clearpage
\endgroup

\end{document}